\documentclass[11pt]{article}

\usepackage[utf8]{inputenc}
\usepackage[T1]{fontenc}
\usepackage{times}
\usepackage{amsmath,amssymb,amsthm}
\usepackage{graphicx}
\usepackage{booktabs}
\usepackage{colortbl}
\usepackage{xcolor}
\usepackage{hyperref}
\usepackage{cleveref}
\usepackage{subcaption}
\usepackage{natbib}
\usepackage{microtype}
\usepackage{multirow}
\usepackage{array}
\usepackage{enumitem}

\usepackage[
  top=1in, bottom=1in,
  left=1.2in, right=1.2in,
  headheight=14pt
]{geometry}

\newcommand{\logit}{\text{logit}}

\hypersetup{
  colorlinks=true,
  linkcolor=blue!70!black,
  citecolor=blue!70!black,
  urlcolor=blue!70!black
}

\definecolor{streamingblue}{rgb}{0.90, 0.94, 1.0}

\title{%
  \textbf{How Well Can Your Video Model Remember?}\\[4pt]
  Measuring Memory-Budget Trade-offs in Long Video Understanding
}

\author{%
  Yixian Tian$^{1,2}$ \\[6pt]
  {\small $^1$Fudan University \quad
  $^2$Beijing University of Posts and Telecommunications} \\[3pt]
  {\small \texttt{tianyixian.nanshan@foxmail.com}}
}

\date{}

\begin{document}

\maketitle

\begin{abstract}
We introduce a compact empirical model that quantifies how answer accuracy
degrades as a function of frame budget $B$ and temporal distance $D$ in long video
understanding---the question of how a model performs when it must recall content
from $D$ seconds in the past using only a fraction $B$ of total frames.
Long-form video models operate under strict frame budgets, yet no prior framework
predicts how accuracy degrades as $B$ shrinks and the relevant event recedes in time.
We fit a weighted least-squares model on ${\sim}155{,}000$ binary predictions across ten models
and three sampling strategies, deriving a compact law in which logit-accuracy scales
linearly in log-budget with a distance-dependent exponent that itself decays
log-linearly with distance.
We call this exponent the \emph{budget exponent} $\alpha(D)$---the marginal
value of extra frames at distance $D$.
Its intercept captures overall budget responsiveness, while its slope (the
\emph{distance decay rate}) measures how fast this responsiveness degrades
with temporal distance.
The law achieves cell-level weighted $R^2 = 0.05$--$0.75$ across models
(highest for models with strong budget sensitivity, near-zero for models
where budget has minimal effect on accuracy).
Our most striking finding: \textbf{budget effectiveness at $D = 1000$\,s differs
by ${\approx}7.4\times$ between the best streaming model and the best base model}.
\textsc{StreamingVLM} achieves $\alpha(1000) = 1.26$ (95\% CI via video-level
cluster bootstrap: $[1.06, 1.58]$), meaning each tenfold budget increase
substantially increases accuracy even at long distances, while the best Qwen3-VL
base model reaches only $\alpha(1000) = 0.17$ (CI: $[0.04, 0.34]$);
the intervals do not overlap.
In accuracy space, a 10$\times$ budget increase at $D = 1000$\,s yields $+29$
percentage points for \textsc{StreamingVLM} versus $+4$\,pp for the best base model.
Frame sampling strategy has model-dependent effects: random sampling yields higher
base sensitivity $a$ but steeper distance decay $e$, with the optimal choice
depending on the target temporal distance.
We demonstrate how $\alpha(D)$ enables principled budget allocation, including
a model-ranking reversal at long distance that a simpler law would miss,
and propose it as a standardized diagnostic metric for evaluating streaming video models.

\end{abstract}

\section{Introduction}
\label{sec:intro}
Long-form video models must answer questions about events that occurred minutes or
hours before the query.
They do so under a strict \emph{frame budget}: only a fraction $B \in (0,1]$ of video
frames can be processed.
Yet no quantitative framework characterizes \emph{how} answer accuracy degrades as $B$ shrinks
and the relevant event recedes in time by distance $D$.

This gap matters in practice.
As video understanding systems are deployed on hour-long
recordings---security footage, surgical video, lecture
archives---engineers must choose how many frames to process and
which architectural properties ensure that distant events remain recoverable.
Without a predictive framework, these choices are made by intuition rather than by theory.

\textbf{Prior work does not provide this framework.}
Scaling laws for language models~\citep{kaplan2020scaling,hoffmann2022training} characterize
training-time resource allocation but say nothing about inference-time frame budgets.
Streaming video architectures~\citep{streamingvlm,livecc,flashvstream} are evaluated on average
accuracy across video lengths but are never compared on the same principled metric
of \emph{budget effectiveness at distance}.
Prior work studies frame selection, token compression, and streaming adaptation under
compute or memory budgets~\citep{gridprobe,cords,fluxmem,r3streaming}, but does not
model accuracy jointly as a function of frame budget and temporal distance.
Long-video benchmarks~\citep{riverbench,videomme,egoschema} aggregate performance over conditions
without systematically varying both $B$ and $D$.

\textbf{Our contribution.}
We derive, to our knowledge, the first compact empirical model that jointly links answer accuracy, frame budget, and
temporal distance across multiple streaming video models:
\begin{equation}
\logit(P) = \alpha(D)\cdot\log_{10} B + \beta\cdot\log_{10} D + \delta\cdot\log_{10} T,
\quad \alpha(D) = a + e\cdot\log_{10} D.
\label{eq:law}
\end{equation}
Here $\alpha(D) = \partial\logit(P)/\partial\log_{10} B$ is the \emph{budget exponent}---the
logit-accuracy gain per tenfold increase in frames at distance $D$.
The base sensitivity $a$ captures how much a model benefits from additional frames overall,
while the \emph{distance decay rate} $e = \partial\alpha/\partial\log D$
measures how fast this benefit degrades as $D$ grows.
A model with high $a$ and $e \approx 0$ is a true streaming model---it extracts
large, distance-invariant value from each additional frame.

\textbf{Overview.}
We fit Equation~\eqref{eq:law} via weighted least squares on ${\sim}155{,}000$ binary predictions
across ten models, 15 budget fractions, and 4 temporal-distance tiers (22s--2400s)
on the RIVER-bench Retro-Memory subset~\citep{riverbench}.
The law achieves cell-level weighted $R^2 = 0.05$--$0.75$ across models
(interpretable fits for models with budget sensitivity; near-zero for budget-insensitive models).

\begin{figure}[!t]
\centering
\begin{subfigure}[t]{\linewidth}
\centering
\includegraphics[width=0.9\linewidth]{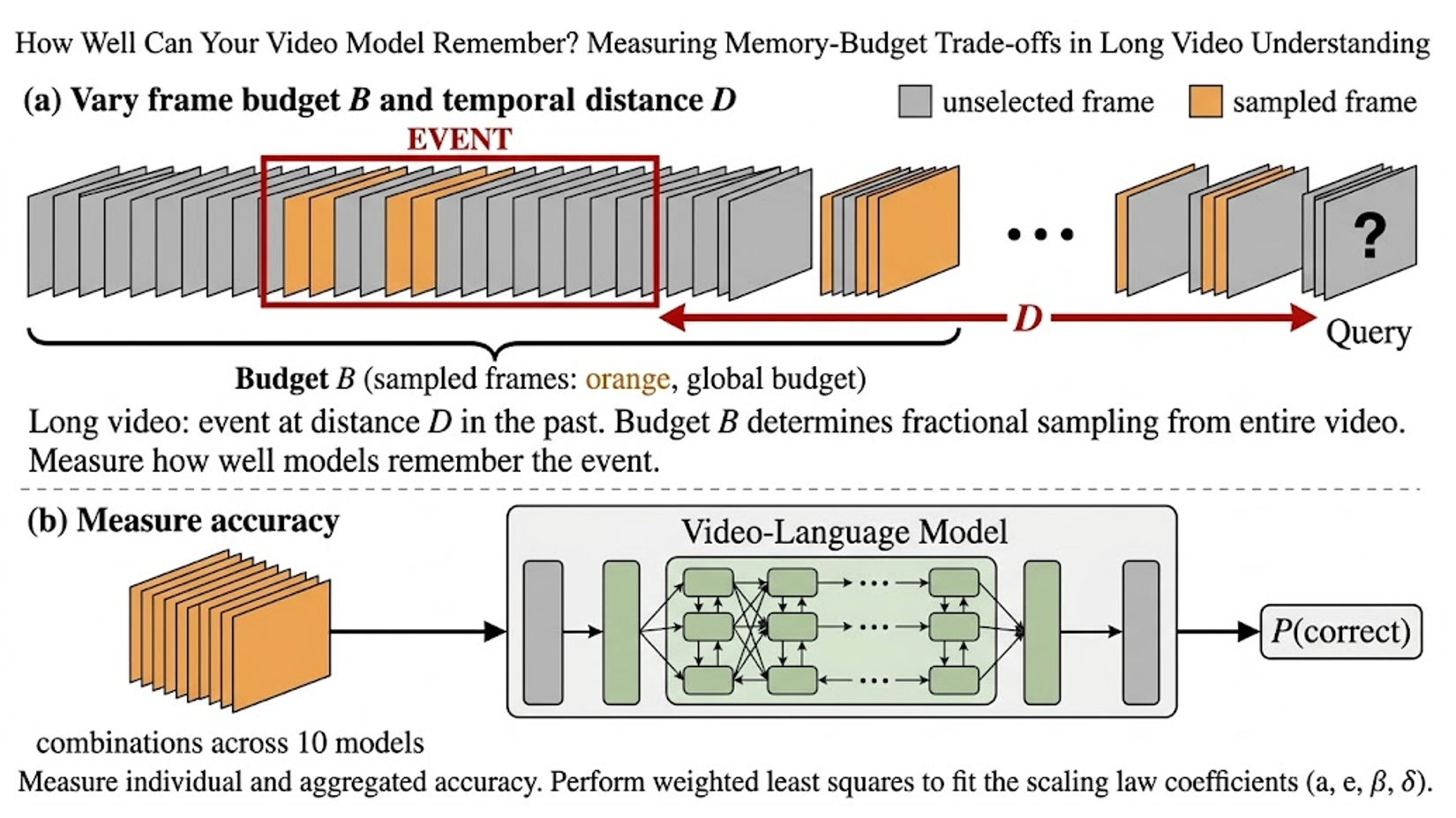}
\caption{For each question, we sample a fraction $B$ of frames from a long video
in which the relevant event occurred at temporal distance~$D$ before the query.
At large~$D$, the same budget covers fewer event-relevant frames.}
\label{fig:hero_concept}
\end{subfigure}

\vspace{4pt}

\begin{subfigure}[t]{\linewidth}
\centering
\includegraphics[width=0.58\linewidth]{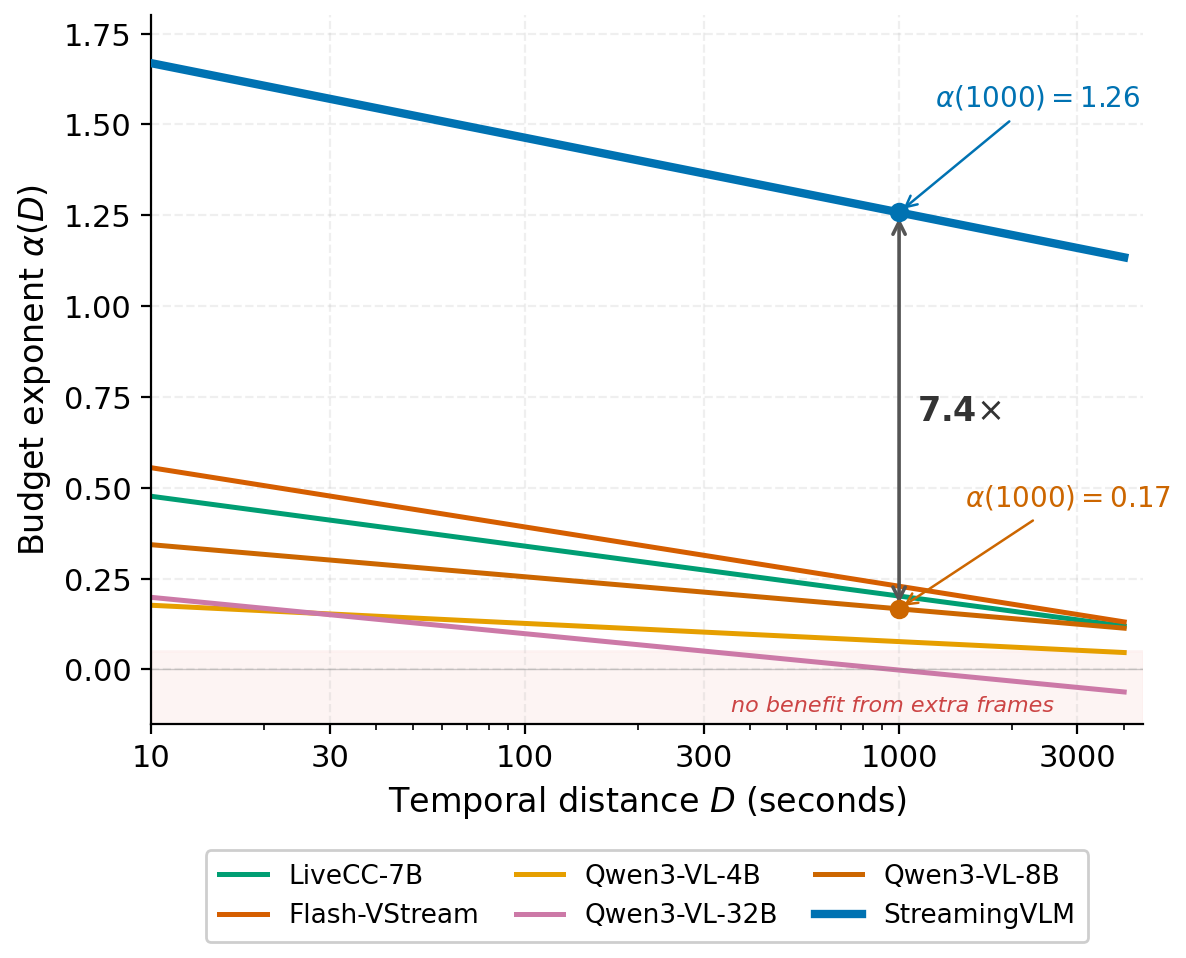}
\caption{The budget exponent $\alpha(D)$ fitted from our scaling law ranks models
on streaming capability.
At $D = 1000$\,s, \textsc{StreamingVLM} maintains $\alpha = 1.26$ while
\textsc{Qwen3-VL-8B} drops to $\alpha = 0.17$---a gap of ${\approx}7.4\times$.}
\label{fig:hero_alpha}
\end{subfigure}
\caption{
\textbf{How we measure budget effectiveness and what it reveals.}
(a)~Experimental setup: varying frame budget~$B$ at different temporal distances~$D$.
(b)~The resulting $\alpha(D)$ curves reveal large differences in how well models
leverage additional frames at long distances.
}
\label{fig:hero}
\end{figure}

\textbf{Contributions.}
\begin{enumerate}
\item
We derive the empirical model (Eq.~\ref{eq:law}), validated with cell-level weighted
$R^2 = 0.05$--$0.75$ across ten models on ${\sim}155{,}000$ predictions, and show the budget exponent $\alpha(D)$ provides
a principled diagnostic metric for comparing streaming capability.

\item
We show that $\alpha(D)$ separates \textsc{StreamingVLM} from the best base model
by ${\approx}7.4\times$ at $D = 1000$\,s:
\textsc{StreamingVLM} achieves $\alpha(1000) = 1.26$ (95\% CI: $[1.06, 1.58]$,
video-level cluster bootstrap) while \textsc{Qwen3-VL-8B}
(the strongest base model at this distance) reaches $\alpha(1000) = 0.17$
(95\% CI: $[0.04, 0.34]$); the intervals do not overlap.
The next-strongest model overall is \textsc{Flash-VStream} (uniform)
at $\alpha(1000) = 0.30$, still a $4.2\times$ gap versus \textsc{StreamingVLM}.

\item
We show that \emph{frame sampling strategy has model-dependent effects}:
random sampling yields higher base budget sensitivity $a$ but steeper distance decay $e$,
with the net effect on long-distance performance varying across architectures.

\item
We illustrate how $\alpha(D)$ enables principled budget allocation and show that
the interaction term $e$ produces model-ranking reversals at long distance that
a simpler distance-independent law would miss.
\end{enumerate}

\textbf{Positioning.}
This paper is a diagnostic study, not a new model.
The contribution is a \emph{standardized measurement protocol}: a compact empirical
model and its derived metric $\alpha(D)$ that allows systematic comparison of how
different architectures handle the budget--distance trade-off.
We show that models already labeled ``streaming'' by their authors differ by $7.4\times$
in budget effectiveness at long distances, and that a simple measurement protocol---fitting
five coefficients per model---can expose these differences where aggregate accuracy cannot.

\section{Related Work}
\label{sec:related}
\paragraph{Scaling laws for training-time resources.}
Kaplan et al.~\citeyearpar{kaplan2020scaling} established that language-model loss follows
a power law in parameters, data, and compute.
Hoffmann et al.~\citeyearpar{hoffmann2022training} (Chinchilla) refined optimal compute allocation.
Analogues exist for vision \citep{zhai2022scaling} and multimodal models \citep{aghajanyan2023scaling}.
All of these characterize \emph{training-time} resource allocation: how much data or compute
to invest to reach a target loss.
\textsc{Quicksviewer}~\citep{quicksviewer} observes a power law between input frame count
and model capability at inference time, motivating their nonuniform video compression method.
However, this observation is incidental to their engineering contribution and does not model
temporal distance as a variable or formalize the budget--accuracy relationship.
Our law addresses an orthogonal problem: at \emph{inference time}, given a fixed model,
how does accuracy degrade jointly as frame budget $B$ shrinks and temporal distance $D$ grows?

\paragraph{Streaming and online video models.}
\textsc{StreamingVLM}~\citep{streamingvlm} processes streaming video by maintaining
a compact KV cache with attention sinks and sliding vision windows.
\textsc{LiveCC}~\citep{livecc} and \textsc{Flash-VStream}~\citep{flashvstream} adapt
Qwen2-VL~\citep{qwen2vl} for online caption generation.
\textsc{VideoLLM-online}~\citep{videollmonline} uses a streaming LoRA adapter on LLaMA.
More recently, \textsc{R3-Streaming}~\citep{r3streaming} introduces agentic
remember-respond-reason control with age-aware forgetting under token budgets,
\textsc{EvoStreaming}~\citep{evostreaming} self-evolves offline models into streaming
assistants via interaction-policy learning,
\textsc{StreamPro}~\citep{streampro} benchmarks proactive streaming responses under
partial observations, and
Yao et al.~\citep{streamingharness} propose a plug-and-play streaming harness
with 12-hour memory retention and sub-second latency.
Prior evaluations compare these models by average accuracy on standard benchmarks,
which does not separate the effect of temporal distance or frame budget.
In our offline retrospective evaluation (all models receive sampled frames via full attention),
\textsc{LiveCC} and \textsc{Flash-VStream} achieve $\alpha(1000) \leq 0.23$ under random sampling
($\leq 0.30$ under uniform)---comparable
to standard base models---despite their streaming-oriented training.
\textsc{StreamingVLM}~\citep{streamingvlm} achieves $\alpha(1000) = 1.26$ ($a = 1.87$)
with random sampling in our offline retrospective evaluation; with recency sampling
(matching its streaming training regime), $\alpha$ rises to $1.71$ due to the
retrospective-task interaction (Section~\ref{sec:results_main}).
\textsc{VideoLLM-online}~\citep{videollmonline} uses a related streaming LoRA adapter
on LLaMA; we do not evaluate it directly but note its architectural similarity
to \textsc{StreamingVLM}.

\paragraph{Long-video benchmarks.}
EgoSchema~\citep{egoschema} provides 5-minute egocentric videos with temporal questions.
Video-MME~\citep{videomme} spans seconds to hours across diverse domains.
MLVU~\citep{mlvu} targets multi-task long-video understanding.
RIVER-bench~\citep{riverbench} is the only benchmark that explicitly structures questions
by temporal distance to the answer, making it suitable for our study.
Neptune~\citep{neptune} exposes long-horizon failure modes but does not vary frame budget.
\textsc{ExtremeWhenBench}~\citep{extremewhenbench} demonstrates that hour-scale
temporal grounding is fundamentally a search problem: all tested Video-LLMs collapse,
while frame-level retrieval baselines succeed---consistent with our finding that
budget becomes critical at long temporal distances.
\textsc{Egostream}~\citep{egostream} introduces a diagnostic benchmark for streaming
episodic memory with an ``Answer Validity Window'' that controls recall distance,
comparing KV-cache management strategies within a unified Qwen3-VL backbone.
Their key finding---that comparable aggregate accuracies mask starkly different
memory profiles---is conceptually aligned with our $\alpha(D)$ diagnostic.
However, \textsc{Egostream} evaluates \emph{KV-cache policies} (pruning, merging,
offloading) in an egocentric streaming setting, while we evaluate
\emph{model architectures} under controlled frame-budget variation in an offline
retrospective setting, and provide a continuous parametric model rather than
per-condition accuracy tables.
Unlike these benchmarks, our analysis is not just a fixed evaluation: it systematically sweeps
both budget \emph{and} distance to derive a predictive law.

\paragraph{Frame sampling and token budgets.}
Uniform sampling---selecting evenly spaced frames---is the default in virtually all video
understanding systems~\citep{livecc,flashvstream,videomme}.
Adaptive or keyframe-based sampling~\citep{focus} requires content
understanding before sampling, adding latency.
\textsc{Swift Sampling}~\citep{swiftsampling} identifies high-information frames via
Taylor-expansion-based temporal surprise detection, achieving +12.5\,pp on long videos
under limited budgets---addressing \emph{which} frames to select, complementary to
our measurement of \emph{how much} budget matters.
\textsc{AdaCodec}~\citep{adacodec} uses predictive visual coding to surpass a
224K-token baseline at $1/7$ the budget on Qwen3-VL-8B, demonstrating that intelligent
frame encoding can invert the budget--accuracy relationship.
\textsc{Q-Fold}~\citep{qfold} constructs query-aware focus--context representations
under fixed visual budgets, gaining up to 9.1\,pp on ultra-long videos.
\textsc{PEEK}~\citep{peek} distills caption-conditioned frame relevance into a
lightweight selector, outperforming uniform sampling especially at 1--2 frames.
\textsc{DynaTok}~\citep{dynatok} retains 95\% of baseline accuracy with 90\% token
reduction via temporally adaptive budget allocation.
\textsc{MuKV}~\citep{mukv} compresses KV caches at patch, frame, and segment
granularities for long streaming video QA.
\textsc{Moment-Video}~\citep{momentvideo} diagnoses temporal fidelity and finds that
denser sampling helps some models but not others---consistent with our finding that
$\alpha$ varies widely across architectures.
\textsc{InfiniPot-V}~\citep{infinipotv} and \textsc{StreamMem}~\citep{streammem}
study memory-efficient architectures under token budgets.
\textsc{GridProbe}~\citep{gridprobe} adaptively allocates test-time compute by probing
frame-importance posteriors for long-video VLMs.
At the architectural level, \textsc{FluxMem}~\citep{fluxmem} and
\textsc{CoRDS}~\citep{cords} compress streaming KV caches via adaptive hierarchical
memory and coreset selection respectively, both operating under fixed memory budgets.
Qu et al.~\citep{lsdbench} introduce LSDBench to study the trade-off between
temporal coverage and spatial detail fidelity under a fixed frame budget.
All of these methods optimize \emph{how} to allocate a budget; our contribution is
a diagnostic \emph{metric} ($\alpha(D)$) that quantifies how much a given model
benefits from budget at each temporal distance---a measurement tool that can
guide the deployment of any budget-allocation strategy.

\section{Empirical Model: Derivation and Validation}
\label{sec:method}
\subsection{Problem Formulation}
\label{sec:formulation}

Consider a video of total length $T$ seconds (ranging from 60\,s to 3600\,s in our experiments).
A question $Q$ is posed at query time $t_q$ (where $t_q \leq T$), and its answer lies in a temporal window
$[t_\text{start}, t_\text{end}]$ with $t_\text{end} < t_q$.
We define \emph{temporal distance} $D = t_q - t_\text{end}$ (seconds; always $D < T$) and
\emph{frame budget} $B \in (0,1]$ (fraction of total frames sampled from $[0, t_q]$).
The outcome $Y \in \{0,1\}$ indicates whether the model answers correctly.

We aim to characterize $P(Y=1)$ as a function of $B$, $D$, and $T$---an \emph{empirical law}
for memory-budget trade-offs on the RIVER-bench Retro-Memory subset.

\paragraph{Experimental setup.}
We use the Retro-Memory subset of RIVER-bench~\citep{riverbench}, which structures
questions by temporal distance to the answer.
\begin{itemize}[nosep]
\item \textbf{Distance tiers}: short ($\bar{D} \approx 23$\,s), medium ($\bar{D} \approx 44$\,s),
      long ($\bar{D} \approx 578$\,s), very long ($\bar{D} \approx 2{,}358$\,s),
      based on the temporal distance from query to answer
\item \textbf{Video length buckets} $T \in \{60, 300, 900, 1800, 3600\}$\,s
\item \textbf{Budget fractions}: 15 levels, $B \in [0.01, 1.0]$
      (see Appendix for full list)
\item \textbf{Models} (10): \textsc{StreamingVLM}~\citep{streamingvlm},
      \textsc{LiveCC-7B}, \textsc{Flash-VStream-7B},
      \textsc{Qwen3-VL-\{2B, 4B, 8B, 32B\}}, \textsc{InternVL3.5-8B},
      \textsc{VideoLLaMA3-7B}, \textsc{VideoLLaMB-7B}
\item \textbf{Sampling strategies}: random (stochastic uniform over $[0, t_q]$),
      uniform (evenly spaced frames), and recency (most recent $B$ frames;
      \textsc{StreamingVLM} only); both random and uniform strategies applied to all models except
      \textsc{Qwen3-VL-2B} (uniform only)
\item \textbf{Scale}: $\approx$155{,}000 binary predictions total across 19 model-strategy combinations
\end{itemize}

\paragraph{Offline retrospective evaluation.}
All models are evaluated in an \emph{offline retrospective} setting: $B$ frames sampled from
$[0, t_q]$ are provided to the model simultaneously as a batch, together with the question.
This is a deliberate design choice: we aim to measure \emph{architectural capacity}
to leverage frames at different temporal distances, independent of any specific
KV-cache management policy.
Real-time streaming evaluation conflates model architecture with cache strategy---a model
may fail because its eviction policy discards relevant tokens, not because its architecture
cannot utilize distant frames.
By providing $B$ frames via full attention, we isolate the architectural factor.
This applies uniformly to all models, including streaming-trained ones
(\textsc{StreamingVLM}, \textsc{LiveCC-7B}, \textsc{Flash-VStream-7B}).
These models were trained with streaming-oriented objectives (e.g., interleaved frame--text
sequences or online caption supervision), but at inference time they process the sampled
frames via standard full attention---the same as base models.
\textsc{StreamingVLM} is additionally evaluated with recency sampling (retaining the $B$
most recent frames), which emulates its streaming KV-cache behavior during training.
The budget $B$ therefore measures \emph{input frame coverage}, not online memory capacity.
This setting isolates the effect of frame budget on retrospective QA accuracy, independent
of any inference-time memory management mechanism.

\subsection{The Empirical Model}
\label{sec:law}

\textbf{Functional form.}
We propose:
\begin{align}
\logit(P) &= \alpha(D)\cdot\log_{10} B + \beta\cdot\log_{10} D + \delta\cdot\log_{10} T + c,
\label{eq:full_law} \\
\text{where}\quad \alpha(D) &= a + e\cdot\log_{10} D.
\label{eq:alpha_D}
\end{align}

\textbf{Motivation.}
The log-log form reflects diminishing returns to budget: each tenfold increase in frames provides
less marginal information at high-budget regimes.
The dependence on $\log D$ reflects the difficulty of longer-term recall.
The interaction term $e\cdot\log B\cdot\log D$ captures that \emph{budget effectiveness
itself changes with distance}---the core novelty of our formulation.

\textbf{Interpretation.}
$\alpha(D) = \partial\logit(P)/\partial\log_{10} B$ is the marginal value of extra frames
at distance $D$.
$e = \partial\alpha(D)/\partial\log_{10} D$ is the \emph{distance decay rate}: how fast this
marginal value degrades as the answer recedes in time.
Note that $\alpha(D)$ measures \emph{budget-distance responsiveness}---how much a model
benefits from additional frames at a given distance---not absolute model quality.
A model with low $\alpha$ may still achieve higher accuracy than a high-$\alpha$ model
at very low budgets, where the high-$\alpha$ model's steep frame dependence becomes a liability.

$e \approx 0$ means that budget and distance are independent---a frame added at any distance
provides the same marginal value (true streaming).
$e \ll 0$ means that at large $D$, budget becomes nearly useless: adding more frames
provides no benefit because the model cannot leverage distant context.

\textbf{Control term.}
The term $\delta\cdot\log_{10} T$ is a control for video length.
The main scaling relation is between $B$, $D$, and $P$; we include $\log T$ to avoid
confounding $D$ effects with video-length effects, not as part of the primary law.

\subsection{Estimation}
\label{sec:estimation}

We estimate the coefficients via \emph{weighted least squares} (WLS) on cell-level
aggregated data.
For each (budget, tier, $T$-bucket) cell with $\geq 5$ predictions and
$B \in [0.03, 1.0)$, we compute the empirical accuracy, clip to $[0.02, 0.98]$,
and apply the logit transform.
The feature vector for each cell is:
\begin{equation}
\mathbf{x} = [\log_{10} B,\; \log_{10} B \cdot \log_{10} D,\; \log_{10} D,\; \log_{10} T,\; 1].
\label{eq:features}
\end{equation}
We solve the weighted normal equations with weights proportional to cell sample size.
The coefficient on $\log_{10} B$ gives $a$, the interaction coefficient gives $e$,
and $\alpha(D) = a + e \cdot \log_{10} D$.
Budget and distance are independently varied across 13 budget levels and 16
(tier, $T$-bucket) combinations (4 tiers $\times$ 5 $T$-buckets less 4 impossible
combinations where $D > T$), yielding up to 208 cells per model-strategy configuration.
This ensures $e$ is identified separately from $a$ and $\beta$.

\textbf{Filtering.}
We exclude $B = 1.0$ (trivial full-budget case) and $B < 0.03$ (extreme low-budget
regime where most cells have too few correct predictions for stable logit estimates).
Distance $D$ is represented by the mean query horizon within each tier.
Bootstrap confidence intervals (Appendix) resample at the cell level only;
they do not account for clustering by video or question, and random-sampling
configurations use a single draw per budget level, so the reported intervals
are likely optimistic.

\subsection{Validation}
\label{sec:validation}

\textbf{In-sample fit.}
The scaling law coefficients reported in Tables~\ref{tab:main_results}
and~\ref{tab:lvo_results} are in-sample fits on the full prediction set per model.
We prioritize in-sample analysis because the law is a \emph{descriptive} tool---it
characterizes the budget--distance--accuracy surface for a given model, not a
predictive model for unseen videos.
We additionally report leave-one-budget-level-out cross-validation in the Appendix
(Table~\ref{tab:held_out_cv}) to assess interpolation across budget levels.

\textbf{Goodness-of-fit.}
We validate the law at the cell level via weighted $R^2$ on aggregated
(budget, distance, $T$) cells, yielding $R^2 = 0.05$--$0.75$ across all
model-strategy configurations.
Models with strong budget sensitivity (\textsc{StreamingVLM}: $R^2 = 0.75$) are well-fit;
models with flat accuracy across budgets (\textsc{VideoLLaMA3-7B}: $R^2 = 0.05$) show
near-zero fit, indicating no systematic budget--distance structure to capture.

\textbf{Interpretability criteria.}
We consider $\alpha(D)$ a meaningful diagnostic for a given model-strategy configuration
only when: (1)~$R^2 \geq 0.20$ (the law captures at least modest systematic variation),
and (2)~the bootstrap 95\% CI for $a$ excludes zero (budget has a statistically
detectable effect).
Of the 19 configurations evaluated, 13 meet both criteria.
For the remaining 6 (\textsc{VideoLLaMA3-7B}, \textsc{VideoLLaMB-7B}, and select
low-budget-sensitivity configurations), $\alpha(D)$ is not interpretable---the
relevant diagnostic finding is simply that these models exhibit negligible
budget-distance structure on this task.

\textbf{Important caveat.}
The law is a \emph{cell-level} model: it predicts how average accuracy varies with
budget and distance, not whether any single prediction will be correct.
Individual binary outcomes are dominated by question-specific difficulty, which the
law does not model.
An instance-level linear probability model yields $R^2 < 0.08$ across all models---expected
given that budget and distance explain systematic trends, not per-question variance.
We therefore do not report per-instance classification accuracy, as it would be
misleading: the majority-class baseline (always predicting the more frequent outcome)
ranges from $0.54$ to $0.63$ across models, leaving little room for a model that
uses only budget and distance as features.

Table~\ref{tab:ablation} reports cell-level weighted $R^2$ for the full law and two ablations.
The interaction term ($e$) improves $R^2$ by $0.002$--$0.02$ across models;
the distance main effect ($\beta \cdot \log D$) contributes $0.00$--$0.03$,
with the largest effect for \textsc{LiveCC-7B}.
While modest individually, these terms are necessary for the law to capture how
$\alpha(D)$ varies with distance---the core diagnostic quantity.
The bulk of variance is explained by $\log B$ alone, consistent with budget being
the dominant factor in accuracy.

\begin{table}[t]
\centering
\caption{
\textbf{Ablation of formula components.}
Cell-level weighted $R^2$ for the full law vs.\ ablations.
The interaction term contributes modest but consistent improvement;
the $\log D$ term matters most for models with strong distance dependence.
}
\label{tab:ablation}
\setlength{\tabcolsep}{5pt}
\small
\begin{tabular}{lccc}
\toprule
\textbf{Model} & \textbf{Full law} & \textbf{No $e$ term} & \textbf{No $\log D$} \\
\midrule
\textsc{StreamingVLM} (random)   & 0.75 & 0.74 & 0.75 \\
\textsc{LiveCC-7B} (random)      & 0.21 & 0.20 & 0.18 \\
\textsc{Flash-VStream} (random)  & 0.47 & 0.45 & 0.45 \\
\textsc{Qwen3-VL-8B} (random)    & 0.32 & 0.32 & 0.32 \\
\bottomrule
\end{tabular}
\end{table}

\begin{figure}[t]
\centering
\includegraphics[width=0.52\linewidth]{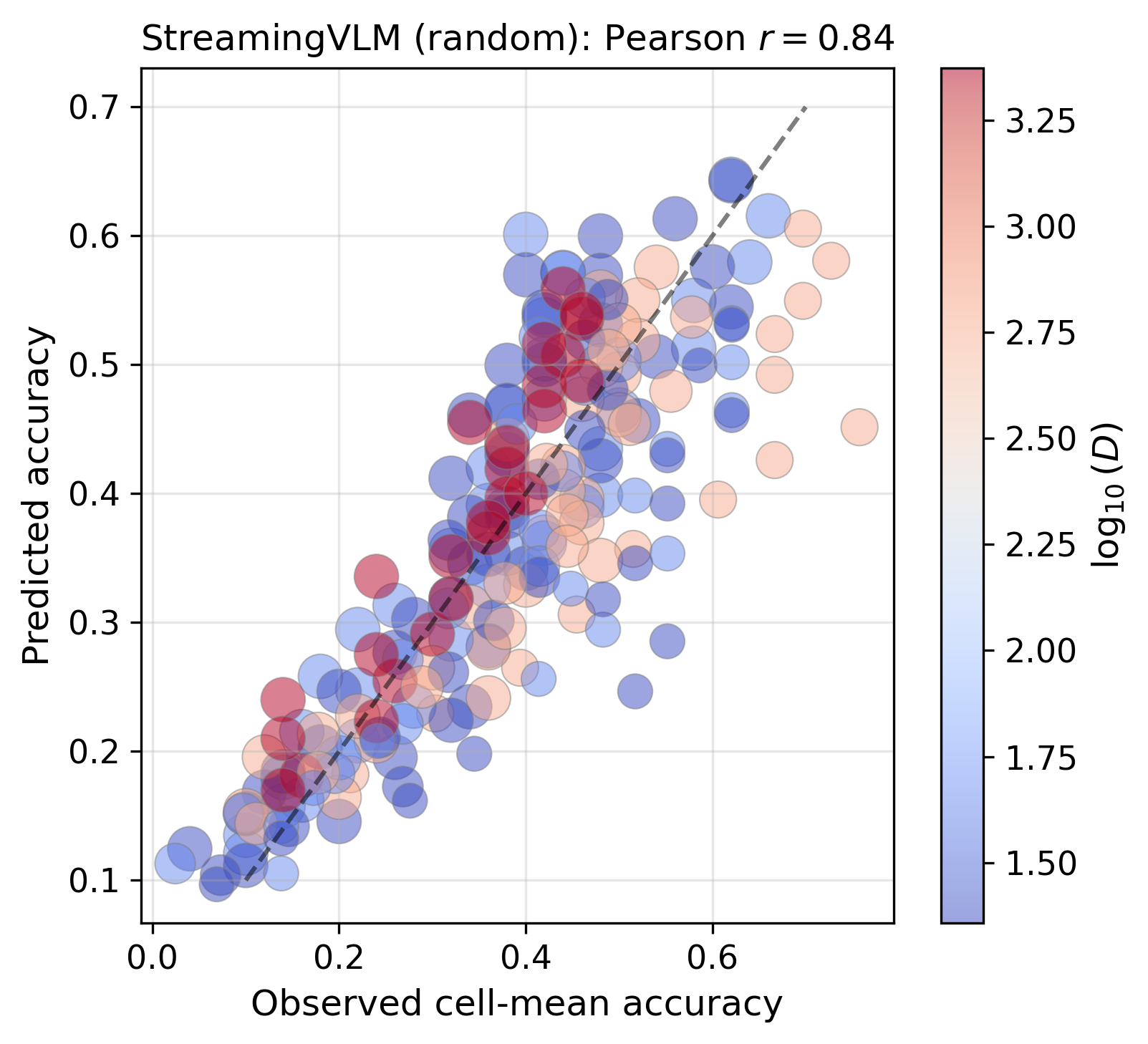}
\caption{
\textbf{Scaling law fit for \textsc{StreamingVLM}.}
Predicted accuracy (from Eq.~\ref{eq:full_law}) versus observed cell-mean accuracy
across all (distance, budget) combinations.
Dot size indicates cell sample count.
}
\label{fig:scatter}
\end{figure}

\textbf{Scatter plot.}
Figure~\ref{fig:scatter} shows predicted versus actual cell-mean accuracy for
\textsc{StreamingVLM} (Pearson $r = 0.84$), the best-fit model.
Across models, Pearson~$r$ ranges from $0.39$ (\textsc{LiveCC-7B})
to~$0.84$ (\textsc{StreamingVLM}), reflecting that the law captures
more structure in models with high budget sensitivity.

\section{Results}
\label{sec:results}
\begin{figure}[t]
\centering
\includegraphics[width=\linewidth]{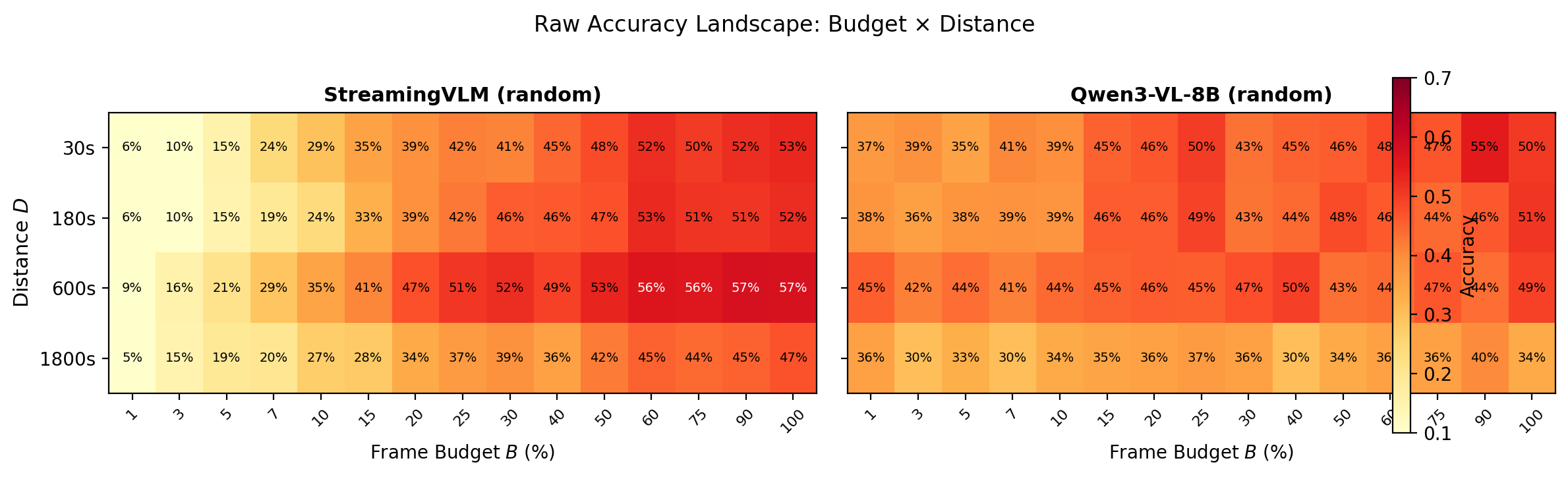}
\caption{
\textbf{Raw accuracy landscape across budget and distance.}
Cell-mean accuracy for \textsc{StreamingVLM} (left) and \textsc{Qwen3-VL-8B} (right),
averaged over video lengths.
\textsc{StreamingVLM} shows a clear gradient along the budget axis at all distances,
while \textsc{Qwen3-VL-8B}'s accuracy is nearly flat---the ``terrain'' that the scaling
law captures as high vs.\ low $\alpha(D)$.
}
\label{fig:accuracy_landscape}
\end{figure}

\subsection{Budget Effectiveness Varies Widely Across Models at Long Distance}
\label{sec:results_main}

Table~\ref{tab:main_results} presents fitted parameters for all models and sampling
strategies.
The law achieves cell-level weighted $R^2$ ranging from $0.05$
(\textsc{VideoLLaMA3-7B}, effectively no fit) to $0.75$
(\textsc{StreamingVLM}, strong fit) across all 19
model-strategy configurations.
Among the 13 configurations with $R^2 \geq 0.20$, the range is $0.21$--$0.75$;
the remaining 6 configurations---\textsc{VideoLLaMA3-7B}, \textsc{VideoLLaMB-7B},
and select \textsc{Qwen3-VL} configurations---exhibit near-flat accuracy across
budget levels, meaning the law cannot capture meaningful structure.
Leave-one-budget-level-out cross-validation shows that out-of-sample $R^2$ drops by
$0.01$--$0.07$ relative to in-sample values across configurations
(Appendix, Table~\ref{tab:held_out_cv}), though 20\% of individual folds yield
negative $R^2$, concentrated at extreme budget levels ($B \leq 0.03$ and $B \geq 0.75$).

\textsc{StreamingVLM} achieves $a = 1.87$ with random sampling, giving
$\alpha(1000) = 1.26$---meaning that at $D = 1000$\,s, each tenfold increase in frame budget
still increases the log-odds of a correct answer by $1.26$ (equivalently, each
doubling adds $1.26 \cdot \log_{10} 2 \approx 0.38$ log-odds).
In contrast, the best Qwen3-VL base model (\textsc{Qwen3-VL-8B}, random) achieves
$\alpha(1000) = 0.17$, a ${\approx}7.4\times$ gap that remains significant under
video-level cluster bootstrap (non-overlapping 95\% CIs).
The next-strongest model overall is \textsc{Flash-VStream} (uniform) at
$\alpha(1000) = 0.30$, a $4.2\times$ gap versus \textsc{StreamingVLM}.

\textbf{Interpreting \textsc{StreamingVLM}'s high $\alpha$.}
\textsc{StreamingVLM} is a streaming-trained model (Qwen2.5-VL-7B fine-tuned with
interleaved frame--text streaming objectives).
Even under standard random sampling at inference time, it retains substantially higher
budget sensitivity at long distances ($\alpha(1000) = 1.26$) compared to all other models
($\alpha(1000) \leq 0.30$).
We additionally evaluate \textsc{StreamingVLM} with \emph{recency} sampling---retaining
the $B$ most recent frames closest to query time $t_q$, matching its training regime---which
yields $\alpha(1000) = 1.71$ (Table~\ref{tab:main_results}), an even steeper budget curve.
However, recency sampling is adversarial for Retro-Memory's retrospective recall task:
at low budget, the frames containing the answer are discarded, driving $B = 0.01$ accuracy
to ${\approx}5\%$ (near chance) while $B = 1.0$ recovers to ${\approx}53\%$.
The inflated $\alpha$ under recency thus reflects the cost of discarding relevant frames,
not solely model capability.
We therefore use the \emph{random-sampling} results as the primary comparison throughout,
as they provide a fairer apples-to-apples evaluation across all models.

\textbf{Confidence intervals.}
Cell-level bootstrap resampling (1000 iterations) yields 95\% CIs of
$[1.48, 2.30]$ for \textsc{StreamingVLM}'s $a$ and
$[0.01, 0.31]$ for \textsc{Qwen3-VL-8B}'s $\alpha(1000)$.
Video-level cluster bootstrap (500 iterations, resampling entire videos to account
for within-video correlation) yields wider but still non-overlapping intervals:
$[1.06, 1.58]$ for \textsc{StreamingVLM}'s $\alpha(1000)$ and $[0.04, 0.34]$ for \textsc{Qwen3-VL-8B}'s.
The gap is robust to both cell-level and video-level resampling variation.
These CIs are conditional on a single random-sampling draw per budget level
and on the fixed cell aggregation.
For some models (e.g., \textsc{LiveCC-7B}), the CI for $e$ includes zero,
indicating that distance decay is not statistically significant---consistent
with the low $R^2$ for these models.

\textbf{Effect sizes in accuracy space.}
To translate $\alpha$ into operational terms, we apply the fitted law
(Eq.~\ref{eq:full_law}) at $D = 1000$\,s and $T = 3600$\,s:
increasing budget from $B = 0.10$ to $B = 1.0$ yields a predicted
accuracy gain of $+29$\,pp for \textsc{StreamingVLM} (from $26\%$ to $55\%$)
versus $+4$\,pp for \textsc{Qwen3-VL-8B} (from $35\%$ to $39\%$).
(These are model-predicted values from Eq.~\ref{eq:full_law}, not raw observations.)
Conversely, cutting budget to $B = 0.01$ drops \textsc{StreamingVLM} to $9\%$
(near chance) while \textsc{Qwen3-VL-8B} remains at $31\%$---still above chance,
but unable to benefit from additional frames.
Importantly, at $B = 0.01$, \textsc{Qwen3-VL-8B} \emph{outperforms}
\textsc{StreamingVLM} in absolute accuracy ($31\%$ vs.\ $9\%$) despite its
much lower $\alpha$.
This illustrates that $\alpha(D)$ measures budget \emph{responsiveness}---not
absolute model quality---and high $\alpha$ can be a liability at very low budgets
where frame dependence means rapid degradation.
The $7.4\times$ gap in $\alpha$ thus corresponds to a $7.3\times$ difference in
accuracy responsiveness to budget ($29$\,pp vs.\ $4$\,pp for a 10$\times$ budget increase).
At short distance ($D = 23$\,s), the picture is reversed in relative terms:
\textsc{StreamingVLM}'s $\alpha(23) = 1.59$ gives $+32$\,pp while \textsc{Qwen3-VL-8B}'s
$\alpha(23) = 0.31$ gives $+8$\,pp---a $4\times$ gap,
confirming that the distance-dependent term $e$ amplifies the separation at long distances.

\textbf{Ratio robustness.}
We report the $\alpha$ ratio as $\approx$7.4$\times$ throughout.
Because \textsc{Qwen3-VL-8B}'s CI lower bound ($0.04$) is near zero,
the bootstrapped ratio distribution is right-skewed.
We therefore also report the absolute gap:
$\Delta\alpha(1000) = 1.26 - 0.17 = 1.09$ (cluster bootstrap 95\% CI: $[0.72, 1.54]$,
derived from the individual-model CIs under independence).
Whether expressed multiplicatively or as an absolute difference, the separation is
substantively large and robust to resampling.

Interestingly, \textsc{LiveCC-7B} and \textsc{Flash-VStream}---despite being designed
for streaming---achieve $\alpha(1000) \approx 0.20$--$0.23$ with random sampling,
closer to base models than to \textsc{StreamingVLM}.
This suggests that the architectural modifications in these models provide limited
benefit for long-distance budget effectiveness.

\begin{table}[t]
\centering
\caption{Scaling law parameters for all models and sampling strategies. $e$ = distance decay rate. $a$ = base budget exponent. $\alpha(D) = a + e\cdot\log_{10} D$ = budget exponent at distance $D$. Ret.\% = $\alpha(1000)/\alpha(10)$, measuring how much budget effectiveness is retained at long distance. $R^2$ = cell-level weighted goodness-of-fit. $\dagger$~95\% CI includes zero. $\ddagger$~Recency row is from a separate evaluation run (Section~\ref{sec:results_main}); included for comparison but not part of the primary analysis. Blue row = primary comparison.}
\label{tab:main_results}
\setlength{\tabcolsep}{3pt}
\small
\begin{tabular}{llcccccc}
\toprule
\textbf{Model} & \textbf{Type} & \textbf{Sampling} & $e$ & $a$ & $\alpha(1000)$ & Ret.\% & $R^2$ \\
\midrule
\texttt{StreamingVLM}$^{\ddagger}$ & Streaming & recency & $-0.09$ & $\mathbf{1.97}$ & $\mathbf{1.71}$ & 91 & 0.85 \\
\rowcolor{blue!7} \texttt{StreamingVLM} & Streaming & random & $-0.21$ & $1.87$ & $1.26$ & 75 & 0.75 \\
\texttt{StreamingVLM} & Streaming & uniform & $-0.17$ & $1.79$ & $1.27$ & 79 & 0.72 \\
\midrule
\texttt{LiveCC-7B} & Streaming & random & $-0.14$ & $0.61$ & $0.20$ & 42 & 0.21 \\
\texttt{LiveCC-7B} & Streaming & uniform & $-0.13$ & $0.58$ & $0.19$ & 43 & 0.23 \\
\texttt{Flash-VStream} & Streaming & random & $-0.16$ & $0.72$ & $0.23$ & 41 & 0.47 \\
\texttt{Flash-VStream} & Streaming & uniform & $-0.09$ & $0.56$ & $0.30$ & 63 & 0.45 \\
\midrule
\texttt{Qwen3-VL-2B} & Base & uniform & $\phantom{-}0.01$ & $0.13$ & $0.15$ & ${\approx}100$ & 0.23 \\
\texttt{Qwen3-VL-4B} & Base & random & $-0.05$ & $0.23$ & $0.08$ & 43 & 0.38 \\
\texttt{Qwen3-VL-4B} & Base & uniform & $-0.20$ & $0.60$ & $0.01$ & 2 & 0.47 \\
\texttt{Qwen3-VL-8B} & Base & random & $-0.09$ & $0.43$ & $0.17$ & 49 & 0.32 \\
\texttt{Qwen3-VL-8B} & Base & uniform & $-0.18$ & $0.60$ & $0.06$ & 15 & 0.27 \\
\texttt{Qwen3-VL-32B} & Base & random & $-0.10$ & $0.30$ & $0.00$ & ${\approx}0$ & 0.31 \\
\texttt{Qwen3-VL-32B} & Base & uniform & $-0.08$ & $0.34$ & $0.10$ & 36 & 0.45 \\
\texttt{InternVL3.5-8B} & Base & random & $-0.54$ & $1.51$ & $-0.12^{\dagger}$ & --- & 0.65 \\
\texttt{InternVL3.5-8B} & Base & uniform & $-0.31$ & $1.13$ & $0.20$ & 25 & 0.71 \\
\texttt{VideoLLaMA3-7B} & Base & random & $\phantom{-}0.05$ & $-0.17^{\dagger}$ & $-0.03^{\dagger}$ & --- & 0.05 \\
\texttt{VideoLLaMA3-7B} & Base & uniform & $-0.02$ & $-0.11^{\dagger}$ & $-0.16^{\dagger}$ & --- & 0.11 \\
\bottomrule
\end{tabular}
\end{table}

\begin{figure}[t]
\centering
\includegraphics[width=0.85\linewidth]{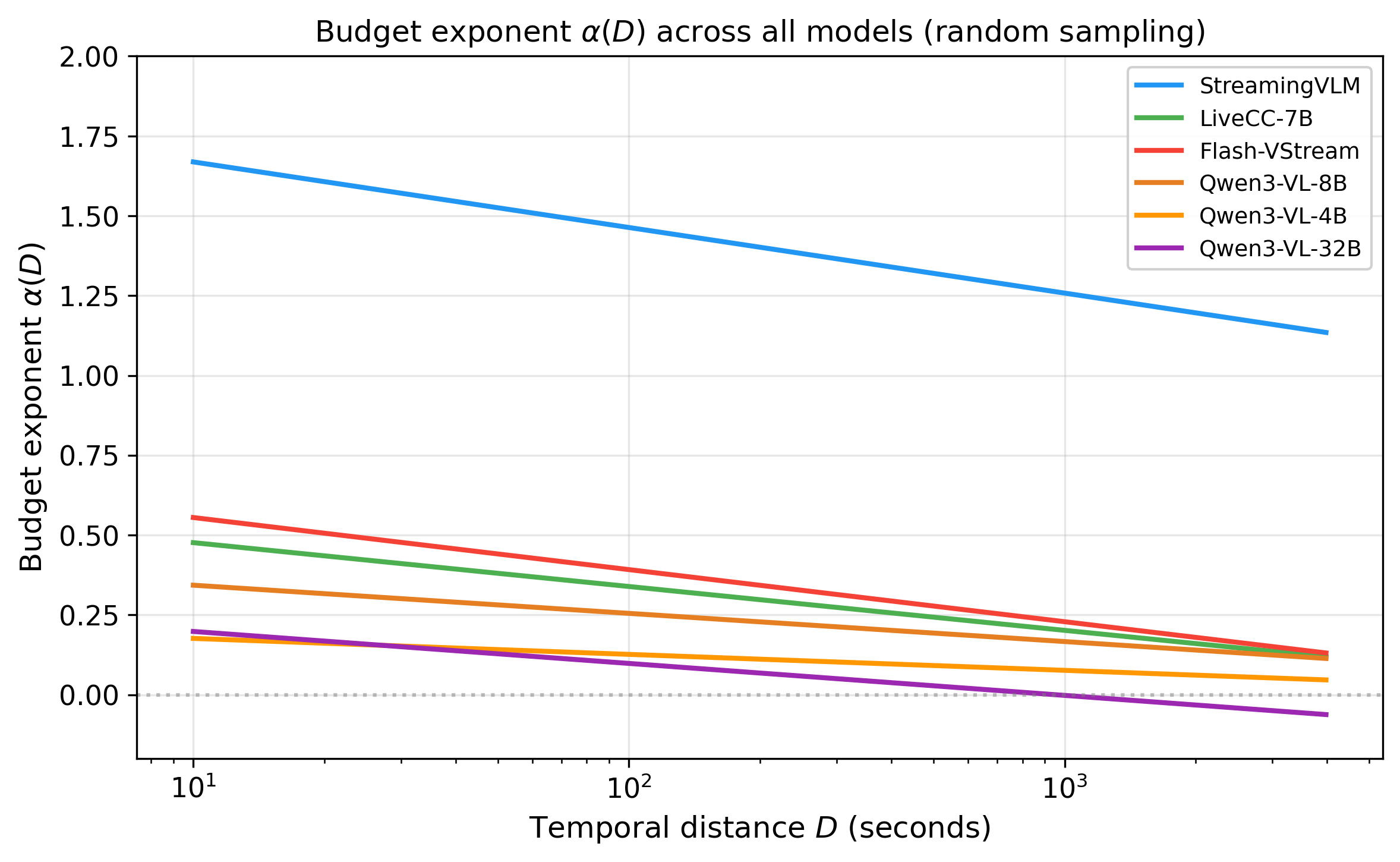}
\caption{
\textbf{Budget exponent $\alpha(D)$ across all models (random sampling).}
\textsc{StreamingVLM} (blue) maintains $\alpha > 1.0$ across all distances,
while base models (orange/purple) and standard streaming models (green/red) remain
below $\alpha = 0.6$.
}
\label{fig:all_models}
\end{figure}

Figure~\ref{fig:all_models} shows $\alpha(D)$ curves for all models.
\textsc{StreamingVLM}'s curve remains above $\alpha = 1.0$ across all distances,
while all other models fall below $\alpha = 0.6$.

\begin{figure}[t]
\centering
\includegraphics[width=0.9\linewidth]{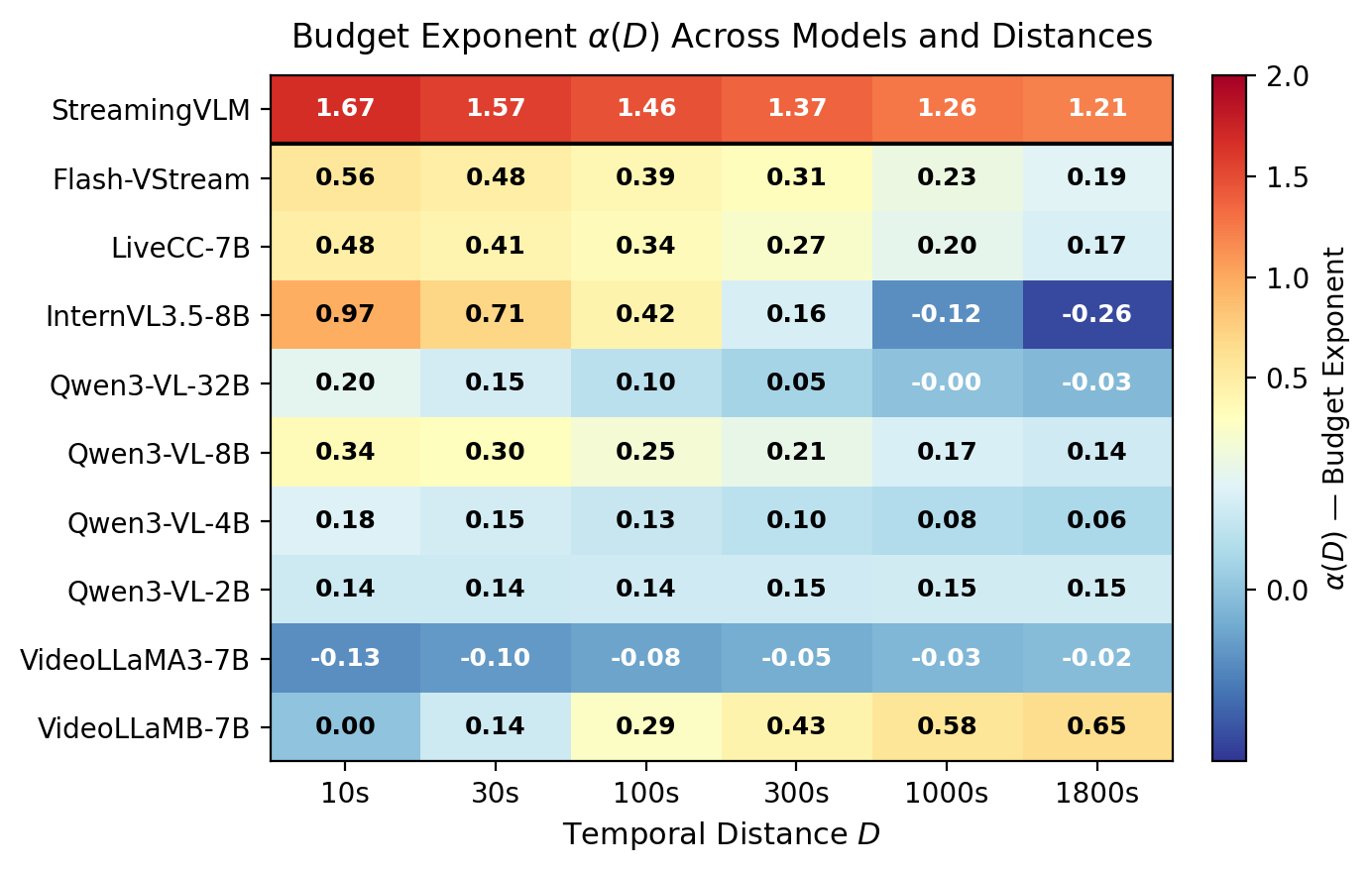}
\caption{
\textbf{Budget exponent $\alpha(D)$ heatmap across models and temporal distances.}
Each cell shows $\alpha(D) = a + e\cdot\log_{10} D$ for a given model at distance $D$
(random sampling; \textsc{Qwen3-VL-2B} uses uniform).
\textsc{StreamingVLM} maintains $\alpha > 1$ everywhere (dark red);
most other models collapse below $0.3$ at $D \geq 300$\,s.
}
\label{fig:alpha_heatmap}
\end{figure}

\subsection{Sampling Strategy Has Model-Dependent Effects}
\label{sec:results_sampling}

Figure~\ref{fig:sampling} compares random versus uniform sampling
for two representative streaming models.
The effect of sampling strategy is more nuanced than a simple ranking:
\begin{itemize}[nosep]
\item Random sampling consistently yields higher base budget sensitivity $a$
for streaming models
(e.g., \textsc{Flash-VStream}: $a = 0.72$ random vs.\ $0.56$ uniform).
\item However, random sampling also produces more negative $e$
(e.g., \textsc{Flash-VStream}: $e = -0.16$ random vs.\ $-0.09$ uniform).
\item The net effect on $\alpha(1000)$ varies: \textsc{Flash-VStream} achieves
$\alpha(1000) = 0.23$ (random) vs.\ $0.30$ (uniform), favoring uniform at long distances.
\end{itemize}

For base models, the pattern is more complex.
\textsc{Qwen3-VL-4B} with uniform sampling achieves higher $a$ ($0.60$ vs.\ $0.23$)
but much steeper decay ($e = -0.20$ vs.\ $-0.05$), resulting in $\alpha(1000) \approx 0.01$
compared to $0.08$ with random sampling.
\textsc{Qwen3-VL-32B} shows the opposite: uniform sampling yields $\alpha(1000) = 0.10$
versus $0.00$ for random.
This confirms that the optimal sampling strategy is model-dependent and cannot be
determined without measuring $\alpha(D)$ at the target distance.

This finding complicates the standard recommendation of uniform sampling:
the optimal strategy depends on the target temporal distance and the specific model
architecture.
For applications requiring long-distance recall, the choice should be guided by
$\alpha(D)$ at the target distance rather than by default convention.

\subsection{Model Scale Alone Does Not Close the Gap}
\label{sec:results_scale}

Scaling \textsc{Qwen3-VL} from 2B to 32B parameters with uniform
sampling yields a non-monotonic pattern in $\alpha(1000)$:
$0.15$ (2B), $0.01$ (4B), $0.06$ (8B), $0.10$ (32B).
The 2B model's relatively high $\alpha(1000) = 0.15$ is noteworthy---it achieves
this through a near-zero interaction term ($e = +0.01$), meaning budget effectiveness
barely decays with distance, despite starting from a low base ($a = 0.13$).
In contrast, the 4B model starts with higher $a = 0.60$ but decays sharply
($e = -0.20$), yielding $\alpha(1000) \approx 0.01$.
This non-monotonic relationship suggests that scale effects on budget effectiveness
are complex and cannot be summarized as a simple capacity threshold within a single
model family.

With random sampling (available for 4B--32B), the 8B model achieves
$\alpha(1000) = 0.17$, the best among the Qwen3-VL family.
In all cases, the Qwen3-VL family remains far below \textsc{StreamingVLM}'s
$\alpha(1000) = 1.26$ (random sampling).
We note that this comparison is observational: the models differ in pretraining data,
architecture, and tuning, so we cannot attribute differences solely to parameter count.

\begin{figure}[t]
\centering
\includegraphics[width=0.72\linewidth]{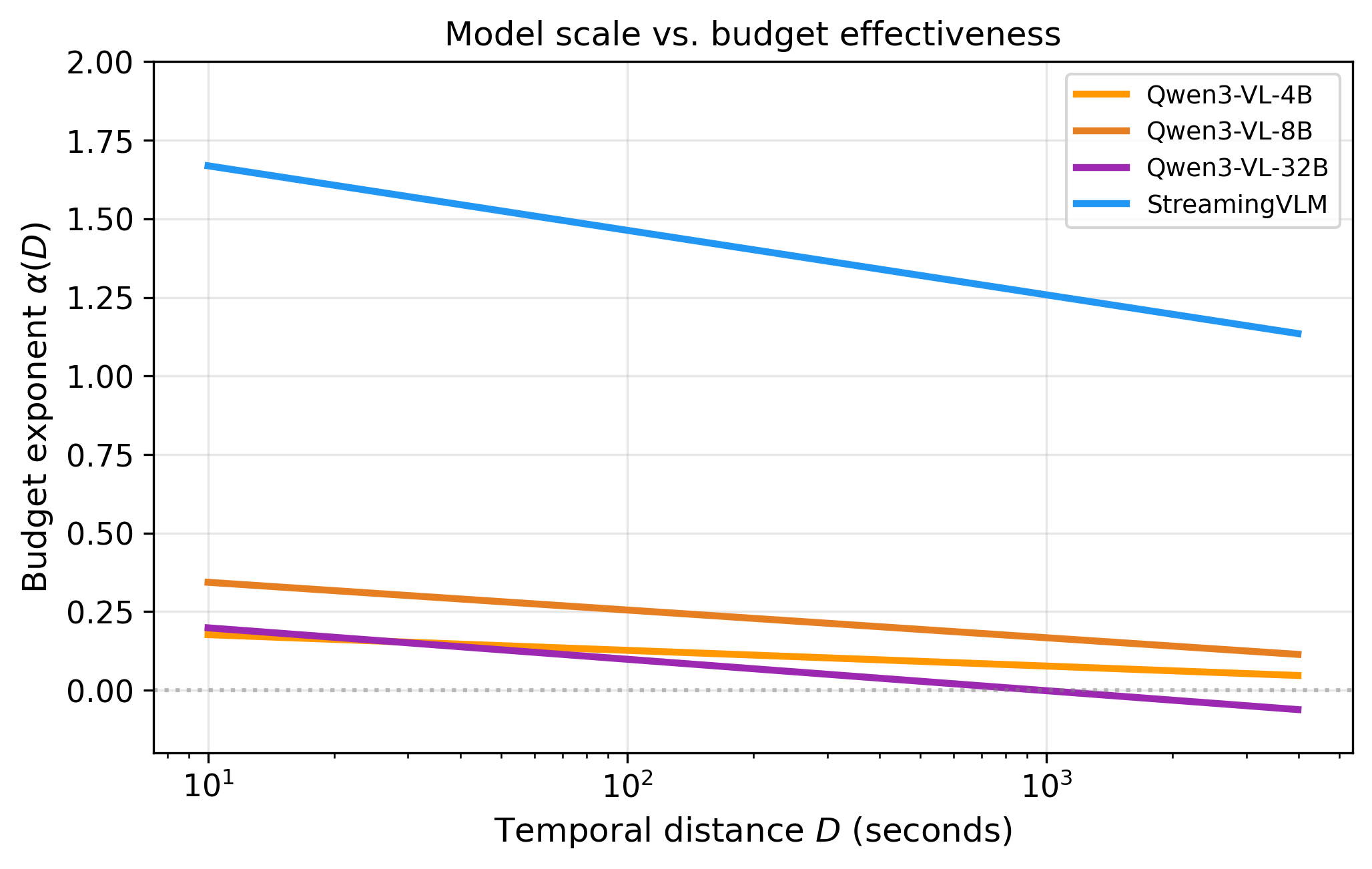}
\caption{
\textbf{Model scale versus budget effectiveness.}
Scaling \textsc{Qwen3-VL} from 4B to 32B parameters does not consistently improve
$\alpha(D)$.
\textsc{StreamingVLM} achieves $\alpha(D) > 1.0$ at all distances through
specialized streaming training.
}
\label{fig:scale}
\end{figure}

Figure~\ref{fig:scale} shows $\alpha(D)$ curves for the Qwen3-VL family alongside
\textsc{StreamingVLM}.

\subsection{Cross-Architecture Evidence}
\label{sec:results_internvl}

To confirm that the budget--distance relationship generalizes beyond the Qwen3-VL family,
we evaluate \textsc{InternVL3.5-8B}, a base model from a different architecture lineage.
\textsc{InternVL3.5-8B} achieves a high base budget sensitivity ($a = 1.51$, random),
between \textsc{StreamingVLM} ($a = 1.87$) and base models---indicating strong budget responsiveness
at short distances.
However, its distance decay rate is the most negative among all models ($e = -0.54$),
meaning budget effectiveness collapses most rapidly with increasing temporal distance.
The result is $\alpha(1000) = -0.12$ (95\% CI: $[-0.52, 0.25]$, includes zero),
indicating that at long distances, additional frames provide no measurable benefit.

This pattern---high $a$ paired with steep $e$---is distinct from the Qwen3-VL models,
which show both lower $a$ and less negative $e$.
\textsc{InternVL3.5-8B} demonstrates that raw model capability (as reflected in high
oracle accuracy of 0.477) does not confer sustained budget sensitivity at long distances.

\textsc{VideoLLaMA3-7B} (based on Qwen2.5-7B) exhibits a third distinct pattern:
$a \approx 0$ with $R^2 < 0.11$ under both sampling strategies, meaning the scaling law
captures essentially no budget--accuracy structure.
All coefficients have 95\% CIs that include zero
($a = -0.17$, CI: $[-0.80, 0.44]$; $\alpha(1000) = -0.03$, CI: $[-0.39, 0.30]$).
Despite achieving above-chance accuracy (${\approx}33\%$ vs.\ $25\%$ random baseline),
\textsc{VideoLLaMA3-7B}'s accuracy is flat across all budget levels---from 2 frames
(budget 5\%) to 48 frames (budget 100\%)---indicating complete budget insensitivity.

\textsc{VideoLLaMB-7B}~\cite{videollamb}, a model with explicit recurrent memory (RMT)
designed for long video understanding, represents a fourth failure mode: answer generation
collapse. Despite its memory-augmented architecture, \textsc{VideoLLaMB-7B} achieves only
$4.4\%$ oracle accuracy on RIVER-bench (below the $25\%$ random baseline for 4-choice MCQ),
with $79\%$ of responses failing to produce a valid answer letter.
The scaling law fit is correspondingly poor ($R^2 = 0.13$--$0.16$), and all coefficients
have wide CIs that include zero.
This result indicates that architectural memory mechanisms alone do not guarantee
temporal memory capability: \textsc{VideoLLaMB-7B}'s RMT design, while effective on its
training benchmarks, does not transfer to RIVER-bench's fine-grained temporal retrieval task.

Together, \textsc{InternVL3.5-8B}, \textsc{VideoLLaMA3-7B}, and \textsc{VideoLLaMB-7B}
illustrate three failure modes for budget effectiveness in base models: rapid distance decay
(high $a$, steep $e$), outright budget insensitivity ($a \approx 0$), and answer generation
collapse (below-chance accuracy).
The finding reinforces the conclusion that high $\alpha(D)$ in streaming-trained models
(\textsc{StreamingVLM}) arises from memory-specific training
rather than from general model capacity,
and that this property is absent in base models across architecture families.
(We note that ``memory'' here refers to budget-distance responsiveness on the
retrospective recall task; see Limitations for the answer-span confound.)

\subsection{Cross-Benchmark Probe on LVOmniBench}
\label{sec:results_lvo}

To probe whether the model \emph{form} (not the specific coefficients) applies
beyond RIVER-bench, we evaluate \textsc{Qwen3-VL} on LVOmniBench~\cite{lvomnibench}, a
long-video QA benchmark with 275 videos (duration 10\,min--60\,min).
For LVOmniBench, the temporal scale proxy $D$ is the video duration in seconds,
since all questions probe content distributed across the full video.
\textbf{Crucially, $D$ here means ``video length'' rather than ``temporal recall distance,''
so coefficients are not directly comparable to RIVER-bench.}
This section demonstrates that the functional form fits a second benchmark,
not that the \emph{metric values} generalize.

We sample up to 15 questions per duration bin at five budget levels
$B \in \{0.05, 0.10, 0.20, 0.50, 1.0\}$, yielding 300 total evaluations per configuration,
and fit the scaling law with $D$ = video duration.
Table~\ref{tab:lvo_results} reports fitted parameters for four configurations:
\textsc{Qwen3-VL-8B} with uniform and random sampling, \textsc{Qwen3-VL-4B} with uniform sampling,
and \textsc{Qwen3-VL-2B} with uniform sampling.

\begin{table}[h]
\centering
\caption{
\textbf{Budget scaling law across benchmarks and model configurations.}
$D$ is temporal recall distance on RIVER-bench and full video duration on LVOmniBench.
The interaction term $e$ has opposite signs across benchmarks ($e < 0$ on RIVER-bench,
$e > 0$ on LVOmniBench for the 8B model), reflecting different semantics of $D$ (see text).
$\alpha(D^\dagger)$: evaluated at $D = 1000$\,s for RIVER-bench, $D = 3600$\,s for LVOmniBench.
The 4B and 2B models yield $e < 0$ and low $R^2$; the poor fit quality
means budget sensitivity cannot be reliably characterized at these model scales.
}
\label{tab:lvo_results}
\small
\begin{tabular}{llccccc}
\toprule
Benchmark & Model & Sampling & $a$ & $e$ & $\alpha(D^\dagger)$ & $R^2$ \\
\midrule
RIVER-bench & \textsc{Qwen3-VL-8B} & uniform & $\phantom{-}0.60$ & $-0.18$ & $0.06$ & $0.27$ \\
RIVER-bench & \textsc{Qwen3-VL-2B} & uniform & $\phantom{-}0.13$ & $\phantom{-}0.01$ & $0.15$ & $0.23$ \\
\midrule
\multirow{4}{*}{LVOmniBench}
 & \textsc{Qwen3-VL-8B} & uniform  & $-1.34$ & $\phantom{-}0.41$ & $\phantom{-}0.12$ & $0.54$ \\
 & \textsc{Qwen3-VL-8B} & random   & $-0.84$ & $\phantom{-}0.24$ & $\phantom{-}0.00$ & $0.37$ \\
 & \textsc{Qwen3-VL-4B} & uniform  & $\phantom{-}1.81$ & $-0.52$ & $-0.03$ & $0.12$\rlap{$^\dag$} \\
 & \textsc{Qwen3-VL-2B} & uniform  & $\phantom{-}1.83$ & $-0.40$ & $\phantom{-}0.42$ & $0.15$\rlap{$^\dag$} \\
\bottomrule
\end{tabular}
\vspace{0.5em}

{\footnotesize $^\dag$ $R^2 \leq 0.15$: fit quality too low for reliable coefficient interpretation.}
\end{table}

\paragraph{Sign of $e$ suggests a capacity effect.}
For \textsc{Qwen3-VL-8B}, the interaction term $e > 0$ under both sampling strategies
($e = +0.41$ uniform, $e = +0.24$ random), indicating that budget sensitivity
\emph{increases} with video duration.
This is the opposite of $e < 0$ on RIVER-bench, and is coherent: video duration captures
\emph{how much content} the model must cover (longer $\Rightarrow$ more frames help),
while temporal recall distance captures \emph{how far back} the model must remember
(longer $\Rightarrow$ relevant frames are diluted across a fixed budget).
By contrast, \textsc{Qwen3-VL-4B} yields $e = -0.52$, $R^2 = 0.12$, and
\textsc{Qwen3-VL-2B} yields $e = -0.40$, $R^2 = 0.15$.
Both sub-8B models show $e < 0$ and low $R^2$ ($\leq 0.15$), but
the low fit quality means the fitted coefficients are not stable enough to
support strong conclusions about budget sensitivity in these models.
We interpret this as preliminary evidence that model capacity interacts with
the ability to systematically leverage additional frames on LVOmniBench,
though the pattern warrants confirmation across model families.

\paragraph{Budget allocation on long videos.}
For \textsc{Qwen3-VL-8B} uniform, $\alpha(3600) = 0.12 > 0$ at the 60-minute tier,
confirming that budget allocation matters for very long videos.
Random sampling yields a weaker effect ($\alpha(3600) \approx 0.00$): higher base sensitivity
($a = -0.84$ vs.\ $-1.34$) is offset by a smaller interaction term ($e = 0.24$ vs.\ $0.41$),
resulting in near-zero budget sensitivity at the longest duration.
Across both benchmarks, the law form captures the budget--temporal-scale structure,
and its parameters encode both benchmark-specific temporal dynamics and model capacity effects.

\subsection{Budget Allocation Application}
\label{sec:results_allocation}

\begin{figure}[t]
\centering
\begin{subfigure}[t]{0.48\linewidth}
\centering
\includegraphics[width=\linewidth]{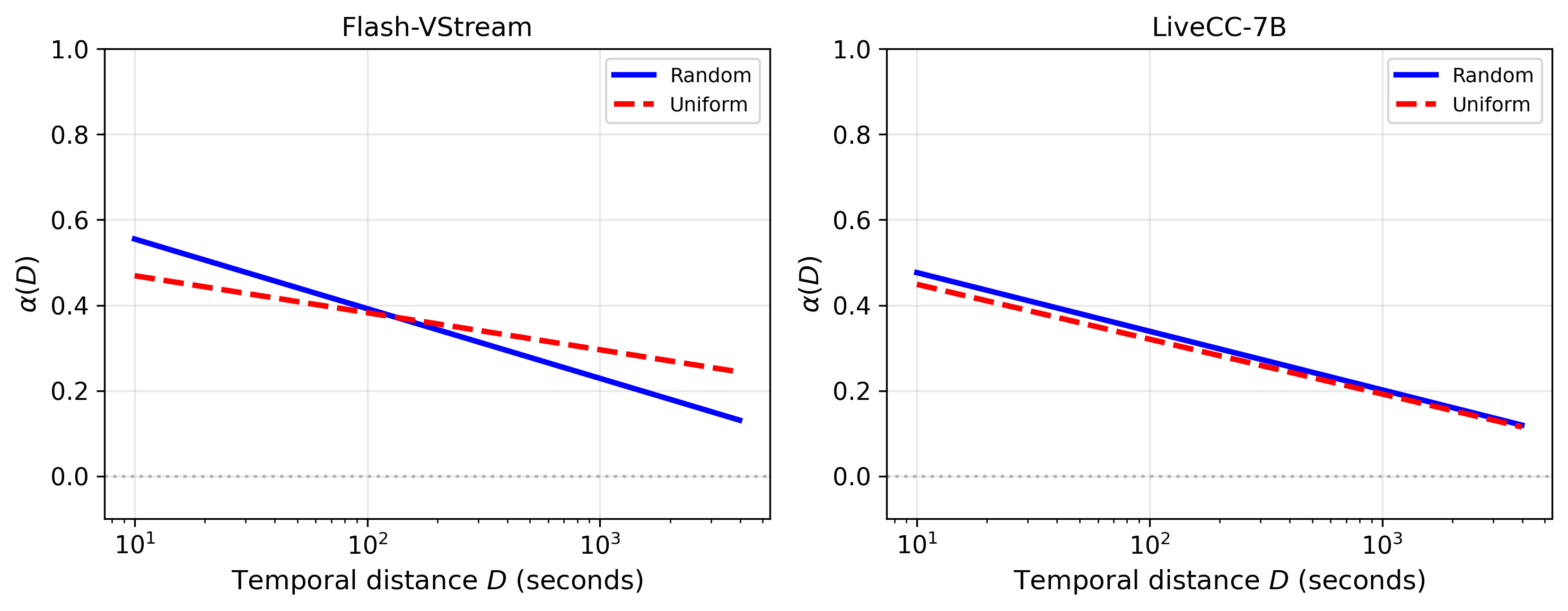}
\caption{Effect of sampling strategy on $\alpha(D)$: random (solid blue) starts higher but decays faster than uniform (dashed red).}
\label{fig:sampling}
\end{subfigure}
\hfill
\begin{subfigure}[t]{0.48\linewidth}
\centering
\includegraphics[width=\linewidth]{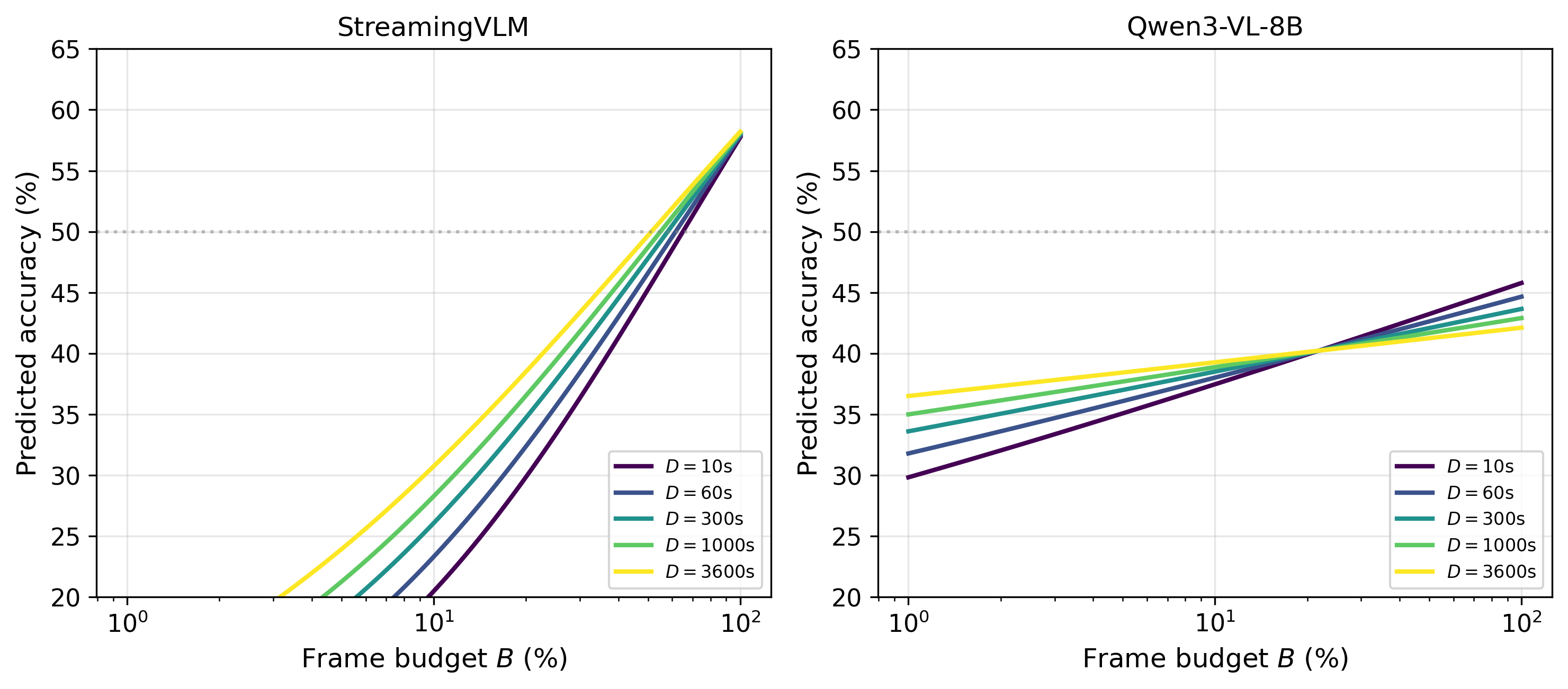}
\caption{Predicted accuracy vs.\ budget at multiple distances for \textsc{StreamingVLM} (left) and \textsc{Qwen3-VL-8B} (right).}
\label{fig:budget}
\end{subfigure}
\caption{
\textbf{Budget sensitivity diagnostics.}
(a)~Sampling strategy affects the $\alpha(D)$ curve shape: random sampling yields higher base sensitivity but steeper distance decay, with model-dependent crossover points.
(b)~The scaling law predicts accuracy as a function of budget at each distance; \textsc{StreamingVLM}'s high $\alpha$ translates budget increases into large accuracy gains even at $D=1000$\,s, while \textsc{Qwen3-VL-8B}'s low $\alpha$ renders additional frames nearly ineffective.
}
\label{fig:budget_diagnostics}
\end{figure}

Given a target accuracy and temporal distance, Equation~\eqref{eq:full_law} predicts
the required frame budget.
Figure~\ref{fig:budget} shows accuracy versus budget at five distances for
\textsc{StreamingVLM} and \textsc{Qwen3-VL-8B}.

\paragraph{The interaction term $e$ produces ranking reversals.}
Without the distance-dependent term ($\alpha = a$, constant), the simpler law
would rank \textsc{Qwen3-VL-32B} above \textsc{Qwen3-VL-4B} in budget effectiveness
at all distances ($a = 0.30$ vs.\ $0.23$, random sampling).
However, the full law with $e$ reverses this ranking at $D = 1000$\,s:
\textsc{Qwen3-VL-4B} achieves $\alpha(1000) = 0.08$ while \textsc{Qwen3-VL-32B}
drops to $\alpha(1000) \approx 0.00$, because the 32B model's steeper decay
($e = -0.10$ vs.\ $-0.05$) erases its short-distance advantage.
A practitioner targeting long-distance recall would choose the wrong model
without the interaction term.

The high $\alpha(D)$ of \textsc{StreamingVLM} means it can achieve target accuracy
levels with far fewer frames:
\begin{itemize}[nosep]
\item At $D = 1000$\,s, \textsc{StreamingVLM}'s $\alpha(1000) = 1.26$ means each
10$\times$ increase in budget raises accuracy substantially.
\item \textsc{Qwen3-VL-8B}'s $\alpha(1000) = 0.17$ means budget increases have
minimal effect at long distances.
\end{itemize}

\section{Discussion}
\label{sec:discussion}
\subsection{Relation to Concurrent Work}

\paragraph{Comparison with \textsc{Egostream}.}
The concurrent \textsc{Egostream} benchmark~\citep{egostream} shares our goal of
diagnosing how streaming models handle temporal recall at varying distances.
However, the two studies differ in three key respects.
First, \textsc{Egostream} evaluates \emph{KV-cache management policies}
(sliding windows, pruning, merging, offloading) applied to a single backbone (Qwen3-VL),
whereas we compare \emph{model architectures} (10 models across 4 families)
under a controlled frame-budget sweep.
Second, \textsc{Egostream} reports per-condition accuracy tables across seven
cognitive dimensions, whereas our $\alpha(D)$ provides a single continuous metric
that enables direct model ranking and budget-allocation predictions.
Third, our offline retrospective evaluation isolates the frame-budget variable
from inference-time cache dynamics, providing a cleaner measurement of
architectural memory capacity.
The two approaches are complementary: \textsc{Egostream} reveals \emph{which}
cache strategy best preserves episodic memory, while our law reveals
\emph{which architectures} benefit from additional frames at long distances.

\paragraph{Distance-aware memory in video generation.}
Independently, \textsc{FadeMem}~\citep{fademem} proposes distance-aware KV memory
consolidation for autoregressive video \emph{generation}, motivated by the observation
that ``fine details decorrelate quickly, while coarse scene structure remains useful
over longer horizons.''
This mirrors our empirical finding that $e < 0$ across all RIVER-bench models:
budget effectiveness decays with temporal distance.
The convergence of this pattern across understanding and generation tasks suggests
that distance-dependent memory decay is a fundamental property of transformer-based
video processing, not an artifact of our specific evaluation setup.

\subsection{Generalizability of the Empirical Model}

Our cross-benchmark evaluation on LVOmniBench provides preliminary evidence that the
scaling law form applies beyond RIVER-bench's temporal memory task, while revealing
that the law's coefficients encode both benchmark-specific temporal dynamics and
model capacity effects.
Importantly, $D$ has different semantics across the two benchmarks---temporal recall
distance on RIVER-bench versus video duration on LVOmniBench---so the sign of $e$
is not directly comparable; rather, the consistent finding is that the law
\emph{form} fits both settings and $\alpha(D)$ remains a useful diagnostic.
For \textsc{Qwen3-VL-8B}, the interaction term $e > 0$ on LVOmniBench under both
sampling strategies ($e = +0.41$ uniform, $e = +0.24$ random)---opposite in sign to all
RIVER-bench models ($e < 0$)---reflects the different semantic meaning of $D$:
video duration captures how much content a model must cover, so longer videos
benefit more from additional frames ($e > 0$); temporal recall distance captures
how far back a model must remember, so budget becomes less effective as the relevant
frames are spread more thinly ($e < 0$).
For \textsc{Qwen3-VL-4B} and \textsc{Qwen3-VL-2B}, the sign reverses
($e = -0.52$ and $e = -0.40$ respectively) and the fit quality is poor
($R^2 = 0.12$ and $0.15$), meaning the fitted coefficients are not stable enough
to support strong conclusions about budget sensitivity in these models---the
law form still fits, but the low $R^2$ suggests the budget--duration relationship
is too noisy to characterize reliably at this model scale.

Leave-one-budget-level-out cross-validation on RIVER-bench
(Table~\ref{tab:held_out_cv}) shows that aggregated out-of-sample $R^2$ drops by
$0.01$--$0.07$ across configurations (median drop: $0.016$),
with overall OOS $R^2$ ranging from $0.16$ (\textsc{Qwen3-VL-2B}) to $0.82$
(\textsc{StreamingVLM}).
However, 20\% of individual folds yield negative $R^2$, concentrated at extreme
budget levels ($B \leq 0.03$ and $B \geq 0.75$), indicating that the law's
predictions are unreliable when extrapolating to very low or very high allocations.
The law is most predictive in the $B \in [0.05, 0.60]$ range.

We view the two benchmarks as complementary in what they reveal about the law.
RIVER-bench isolates the memory-recall signal and shows that architectural design
(not just budget) determines $\alpha$ at long distances.
LVOmniBench shows that the law form applies to general long-video reasoning:
the tested 8B model exhibits $e > 0$ and fitted $e$ and $\alpha(D)$
remain useful diagnostics even when $D$ is a coarse proxy (video duration) rather than
a precisely defined temporal recall distance;
broader model family coverage is needed to confirm whether this pattern generalizes.

\subsection{Implications for Model Design and Deployment}

\paragraph{Optimize for high $\alpha(D)$, not just accuracy.}
A model with $\alpha(1000) > 1.0$ benefits substantially from additional frames even at
long distances; one with $\alpha(1000) \approx 0$ cannot improve at long distances
regardless of budget (within the $B \in (0,1]$ range).
Practitioners should select models by $\alpha(D)$ at the target distance
and allocate budgets via Equation~\eqref{eq:full_law}.

\paragraph{Training paradigm matters more than scale in our experiments.}
\textsc{StreamingVLM}'s $\alpha(1000) = 1.26$ (random sampling)
versus $0.17$ for \textsc{Qwen3-VL-8B} suggests that streaming-oriented training is key,
though this comparison is confounded by pretraining differences.
Under recency sampling (which matches its training regime), \textsc{StreamingVLM}
achieves $\alpha(1000) = 1.71$, but this inflated value reflects the task--sampling
interaction rather than pure model capability (Section~\ref{sec:results_main}).
Standard streaming-trained models achieve $\alpha(1000) \approx 0.20$--$0.23$, closer to
base models than to \textsc{StreamingVLM}.

\paragraph{Sampling strategy involves a trade-off.}
Random sampling yields higher $a$ but steeper $e$; the net effect at a given distance
is model-dependent.
For \textsc{Flash-VStream}, uniform sampling achieves better $\alpha(1000)$
($0.30$ vs.\ $0.23$); for \textsc{StreamingVLM} the strategies are comparable ($1.26$ vs.\ $1.27$).

\subsection{Limitations}

\paragraph{Benchmark and task scope.}
Main results are from RIVER-bench Retro-Memory (factual recall).
We additionally validate on LVOmniBench, a general long-video QA benchmark, where
the 8B model yields $e > 0$ and $\alpha(3600) = 0.12$ at the 60-minute tier,
while the 4B and 2B models yield $e < 0$ and $R^2 \leq 0.15$, with fit quality
too low to draw strong conclusions about budget sensitivity at this scale
(Section~\ref{sec:results_lvo}).
Broader model family coverage would strengthen generalization claims.
Stratified analysis reveals that video length ($T$) and temporal distance ($D$) are
independent dimensions: accuracy varies across both, with \textsc{StreamingVLM} showing
more stable performance on long videos at long distances compared to base models.

\paragraph{Interaction term contribution.}
The interaction $e \cdot \log B \cdot \log D$ improves cell-level $R^2$ by
$0.002$--$0.02$ across models (Table~\ref{tab:ablation}).
While modest in variance-explained terms, this term is necessary for $\alpha(D)$ to
vary with distance---the core diagnostic quantity.
Leave-one-distance-tier-out analysis (Appendix, Table~\ref{tab:tier_sensitivity})
confirms that $\alpha(1000)$ is stable within $\pm 0.08$ when any single tier is
dropped, and the StreamingVLM--base gap exceeds $1.0$ under all conditions.
The small $R^2$ improvement reflects that budget ($\log B$) explains most cell-level
variance; the interaction captures how budget \emph{effectiveness} changes with
distance, which is a second-order but practically important effect.
Without it, the model would predict identical budget sensitivity at all distances,
missing the ranking reversals documented in Section~\ref{sec:results_allocation}.
We therefore retain the interaction as a diagnostic refinement, not as evidence of a
strong distance-budget coupling in all models.

\paragraph{Variable fit quality.}
The law fits well for models with strong budget sensitivity ($R^2 = 0.75$ for
\textsc{StreamingVLM}) but poorly for budget-insensitive models
($R^2 = 0.05$ for \textsc{VideoLLaMA3-7B}, $R^2 = 0.21$ for \textsc{LiveCC-7B}).
This is expected: when a model's accuracy barely changes with budget, there is
little systematic variation for the law to capture.
The law is most informative precisely for models where budget allocation matters---those
with high $\alpha(D)$.
For low-$\alpha$ models, the main takeaway is that budget has minimal effect,
which is itself a useful diagnostic finding.

\paragraph{Answer span as an uncontrolled confound.}
Our formulation defines temporal distance as $D = t_q - t_\text{end}$, treating the
answer as a point in time.
In practice, answers span a temporal window $[t_\text{start}, t_\text{end}]$ of varying
width---a brief activity (e.g., ``someone falls'', $\sim$2\,s) versus a prolonged event
(e.g., ``the meeting discussion'', $\sim$10\,min).
Under uniform or random sampling, the probability of capturing at least one frame from
the answer window scales with $B \cdot (t_\text{end} - t_\text{start}) / t_q$, so
narrow-span activities require substantially higher budget to sample reliably, while
wide-span events are captured even at low budget.
This answer-span width is a confound that our model does not control for: two questions
with the same $D$ but different span widths will exhibit different effective budget
sensitivity.
The high $\alpha$ of \textsc{StreamingVLM} could partly reflect that its streaming
training makes it better at localizing narrow-span events, not purely that it has
better temporal memory.
RIVER-bench does not annotate answer span width, so we cannot stratify directly;
incorporating span width as an additional covariate (or stratifying by question type
as a proxy) is a natural extension that could sharpen the model's predictions and
disentangle memory from localization ability.

\paragraph{Offline retrospective evaluation.}
All models are evaluated in an offline retrospective setting: $B$ frames are provided
simultaneously as a batch, not processed online.
This applies to streaming-trained models (\textsc{StreamingVLM}, \textsc{LiveCC-7B},
\textsc{Flash-VStream-7B}) as well as base models.
\textsc{StreamingVLM} is additionally evaluated with recency sampling (the $B$ most
recent frames), which matches its streaming training regime but is adversarial for
Retro-Memory's retrospective recall task: at low budget, the frames containing the
answer are discarded.
This inflates $\alpha$ relative to random sampling and should be interpreted as a
property of the recency--retrospective interaction, not solely of the model.
The budget $B$ therefore measures input frame coverage rather than online memory capacity.
Evaluating these models in their native streaming inference mode---where the KV cache
window size serves as the budget---is a natural extension that would directly test
inference-time memory management.

\paragraph{Other limitations.}
Our law uses fractional budget $B$ rather than absolute frame count; we include $\log T$
as a control but a full analysis with absolute counts is future work.
We test only random and uniform sampling.
Our WLS estimation treats cell-level observations as independent; clustered standard errors
(by video or question) would give more conservative intervals.
Random sampling uses a single draw per budget level; multiple seeds would provide
confidence intervals on $a$ and $e$.

\section{Conclusion}
\label{sec:conclusion}
We derived a compact empirical model for memory-budget trade-offs in streaming
video understanding, validated on RIVER-bench Retro-Memory with ${\sim}155{,}000$ predictions
across ten models and three sampling strategies, with cross-benchmark evaluation on LVOmniBench.
The budget exponent $\alpha(D) = a + e \cdot \log_{10} D$ differs by ${\approx}7.4\times$ at
$D = 1000$\,s between \textsc{StreamingVLM} ($1.26$, 95\% CI: $[1.06, 1.58]$) and
the best base model ($0.17$, CI: $[0.04, 0.34]$); the intervals do not overlap
under video-level cluster bootstrap.
In accuracy space, this translates to a 29\,pp gain from a 10$\times$ budget increase
for \textsc{StreamingVLM} versus only 4\,pp for the best base model at $D = 1000$\,s.
Leave-one-distance-tier-out analysis confirms that no single tier drives the result
(maximum $|\Delta\alpha(1000)| = 0.08$).
Sampling strategy involves a model-dependent trade-off between base sensitivity $a$
and distance decay $e$.
We propose $\alpha(D)$ as a standardized diagnostic metric for streaming capability on this benchmark
and encourage future work on adaptive sampling, architectural analysis of high-$\alpha$ models,
and validation on additional benchmarks.

\bibliographystyle{abbrvnat}
\bibliography{references}

\appendix

\section{Implementation Details}
\label{app:details}
\subsection{Hyperparameters and Inference Settings}

All models were evaluated with greedy decoding:
\begin{itemize}[nosep]
\item Temperature: 0.0
\item Max output tokens: 512
\item Top-$p$: 1.0
\item No beam search
\end{itemize}

For each model-strategy combination, we evaluate all 15 budget fractions
$\times$ 4 distance tiers $\times$ 5 video-length buckets, yielding
$\approx$10{,}770 binary predictions per combination.
With 17 model-strategy combinations (9 models $\times$ 2 sampling strategies,
except \textsc{Qwen3-VL-2B} with uniform only),
the total dataset comprises $\approx$155{,}000 predictions.
\textsc{StreamingVLM} contributes $\approx$12{,}188 predictions per strategy (random and uniform),
\textsc{InternVL3.5-8B} contributes $\approx$4{,}500 predictions (random and uniform),
and remaining combinations contribute $\approx$10{,}770 each.

\subsection{Hardware Specifications}

Experiments were conducted on a single node with:
\begin{itemize}[nosep]
\item 8$\times$ NVIDIA H200 GPUs (141\,GB HBM3e each)
\item 2.0\,TiB system RAM
\item Intel Xeon Platinum 8558 (192 cores)
\end{itemize}

Each model-strategy combination was evaluated on a single GPU.
Total wall-clock time was approximately 72 hours across all experiments.

\subsection{Scaling Law Fitting Details}

The WLS fitting procedure operates on cell-level aggregated data:
\begin{enumerate}[nosep]
\item For each (budget, tier, $T$-bucket) cell, compute the empirical accuracy
      from all predictions in that cell.
\item Exclude cells with $B \geq 1.0$, $B < 0.03$, or fewer than 5 predictions.
\item Clip accuracy to $[0.02, 0.98]$ and apply the logit transform.
\item Represent distance $D$ by the mean query horizon within each tier.
\item Fit the feature vector $[\log_{10} B,\; \log_{10} B \cdot \log_{10} D,\;
      \log_{10} D,\; \log_{10} T,\; 1]$ via weighted least squares with weights
      proportional to cell sample size.
\end{enumerate}

\section{Extended Results}
\label{app:results}
\subsection{Per-Tier Accuracy Breakdown}

Table~\ref{tab:per_tier} reports the overall accuracy (fraction of correct predictions)
for each model at each distance tier, averaged across all budget fractions.
These raw accuracies complement the scaling law parameters in Table~\ref{tab:main_results}
by showing absolute performance levels.

\begin{table}[h]
\centering
\caption{Overall accuracy by distance tier (random sampling, all budget fractions).}
\label{tab:per_tier}
\setlength{\tabcolsep}{4pt}
\small
\begin{tabular}{lcccc}
\toprule
\textbf{Model} & \textbf{Short} & \textbf{Medium} & \textbf{Long} & \textbf{Very Long} \\
 & ($\bar{D}\!\approx\!23$s) & ($\bar{D}\!\approx\!44$s) & ($\bar{D}\!\approx\!578$s) & ($\bar{D}\!\approx\!2358$s) \\
\midrule
\texttt{StreamingVLM} & 0.32 & 0.33 & 0.35 & 0.31 \\
\texttt{StreamingVLM} (random) & 0.36 & 0.35 & 0.41 & 0.32 \\
\texttt{LiveCC-7B} & 0.48 & 0.48 & 0.48 & 0.35 \\
\texttt{Flash-VStream} & 0.43 & 0.42 & 0.43 & 0.32 \\
\texttt{Qwen3-VL-4B} & 0.46 & 0.43 & 0.44 & 0.32 \\
\texttt{Qwen3-VL-8B} & 0.44 & 0.43 & 0.44 & 0.34 \\
\texttt{Qwen3-VL-32B} & 0.46 & 0.44 & 0.45 & 0.38 \\
\bottomrule
\end{tabular}
\end{table}

\textbf{Note:} These are raw accuracy values averaged across all budget fractions.
The scaling law captures how accuracy \emph{changes} with budget at each distance,
not the absolute accuracy level.
A model with high $\alpha(D)$ but lower raw accuracy (e.g., \textsc{StreamingVLM}
at the very-long tier) benefits more from additional frames than a model with
higher raw accuracy but low $\alpha(D)$.

\subsection{Full Scaling Law Coefficients}

Table~\ref{tab:full_coeffs} reports all five coefficients of the scaling law
for each model-strategy combination.

\begin{table}[h]
\centering
\caption{Full scaling law coefficients for all model-strategy combinations.}
\label{tab:full_coeffs}
\setlength{\tabcolsep}{3pt}
\small
\begin{tabular}{llccccc}
\toprule
\textbf{Model} & \textbf{Sampling} & $a$ & $e$ & $\beta$ & $\delta$ & $c$ \\
\midrule
\texttt{StreamingVLM} & recency & 1.97 & $-0.09$ & $0.12$ & $-0.45$ & $1.60$ \\
\texttt{StreamingVLM} & uniform$^*$ & 2.14 & $-0.17$ & $0.01$ & $-0.31$ & $1.44$ \\
\texttt{StreamingVLM} & random & 1.87 & $-0.21$ & $0.01$ & $-0.26$ & $1.11$ \\
\texttt{StreamingVLM} & uniform & 1.79 & $-0.17$ & $-0.03$ & $-0.21$ & $0.99$ \\
\texttt{LiveCC-7B} & random & 0.61 & $-0.14$ & $-0.20$ & $-0.20$ & $1.06$ \\
\texttt{LiveCC-7B} & uniform & 0.58 & $-0.13$ & $-0.20$ & $-0.25$ & $1.17$ \\
\texttt{Flash-VStream} & random & 0.72 & $-0.16$ & $-0.14$ & $-0.34$ & $1.19$ \\
\texttt{Flash-VStream} & uniform & 0.56 & $-0.09$ & $-0.06$ & $-0.40$ & $1.18$ \\
\texttt{Qwen3-VL-2B} & uniform & 0.13 & $\phantom{-}0.01$ & $-0.20$ & $-0.33$ & $0.76$ \\
\texttt{Qwen3-VL-4B} & random & 0.23 & $-0.05$ & $-0.07$ & $-0.41$ & $1.11$ \\
\texttt{Qwen3-VL-4B} & uniform & 0.60 & $-0.20$ & $-0.18$ & $-0.39$ & $1.31$ \\
\texttt{Qwen3-VL-8B} & random & 0.43 & $-0.09$ & $-0.06$ & $-0.38$ & $1.07$ \\
\texttt{Qwen3-VL-8B} & uniform & 0.60 & $-0.18$ & $-0.16$ & $-0.27$ & $0.91$ \\
\texttt{Qwen3-VL-32B} & random & 0.30 & $-0.10$ & $-0.04$ & $-0.37$ & $0.97$ \\
\texttt{Qwen3-VL-32B} & uniform & 0.34 & $-0.08$ & $-0.09$ & $-0.37$ & $1.07$ \\
\texttt{InternVL3.5-8B} & random & 1.51 & $-0.54$ & $-0.33$ & $-0.84$ & 3.28 \\
\texttt{InternVL3.5-8B} & uniform & 1.13 & $-0.31$ & $-0.21$ & $-0.92$ & 3.26 \\
\texttt{VideoLLaMA3-7B} & random & $-0.17$ & $\phantom{-}0.05$ & $-0.07$ & $-0.23$ & $0.46$ \\
\texttt{VideoLLaMA3-7B} & uniform & $-0.11$ & $-0.02$ & $-0.12$ & $-0.20$ & $0.36$ \\
\bottomrule
\end{tabular}
\end{table}

\subsection{Bootstrap Confidence Intervals}

Table~\ref{tab:bootstrap_ci} reports 95\% bootstrap confidence intervals
(1000 cell-level resamples) for the key coefficients $a$, $e$, and $\alpha(1000)$.

\begin{table}[h]
\centering
\caption{Bootstrap 95\% confidence intervals for key scaling law parameters (random sampling).}
\label{tab:bootstrap_ci}
\setlength{\tabcolsep}{3pt}
\small
\begin{tabular}{lccc}
\toprule
\textbf{Model} & $a$ [95\% CI] & $e$ [95\% CI] & $\alpha(1000)$ [95\% CI] \\
\midrule
\texttt{StreamingVLM} (recency) & 1.97 [1.57, 2.41] & $-0.09$ [$-0.30$, $0.12$] & 1.71 [1.44, 1.99] \\
\texttt{StreamingVLM} (random) & 1.87 [1.48, 2.30] & $-0.21$ [$-0.39$, $-0.04$] & 1.26 [1.08, 1.43] \\
\texttt{LiveCC-7B} & 0.61 [0.24, 0.99] & $-0.14$ [$-0.29$, $0.02$] & 0.20 [0.02, 0.38] \\
\texttt{Flash-VStream} & 0.72 [0.47, 0.99] & $-0.16$ [$-0.29$, $-0.05$] & 0.23 [0.05, 0.36] \\
\texttt{Qwen3-VL-4B} & 0.23 [$-0.04$, 0.53] & $-0.05$ [$-0.21$, $0.09$] & 0.08 [$-0.13$, 0.27] \\
\texttt{Qwen3-VL-8B} & 0.43 [0.15, 0.74] & $-0.09$ [$-0.22$, $0.03$] & 0.17 [0.01, 0.31] \\
\texttt{Qwen3-VL-32B} & 0.30 [0.06, 0.53] & $-0.10$ [$-0.21$, $0.00$] & $-0.00$ [$-0.15$, 0.13] \\
\texttt{InternVL3.5-8B} & 1.51 [0.89, 2.18] & $-0.54$ [$-0.87$, $-0.27$] & $-0.12$ [$-0.52$, 0.25] \\
\texttt{VideoLLaMA3-7B} & $-0.17$ [$-0.80$, 0.44] & $\phantom{-}0.05$ [$-0.24$, $0.33$] & $-0.03$ [$-0.39$, 0.30] \\
\bottomrule
\end{tabular}
\end{table}

\subsection{Held-Out Predictive Validation}

To assess whether the scaling law overfits to in-sample cells,
we perform leave-one-budget-level-out cross-validation:
for each of the 13 budget levels $B \in [0.03, 0.9]$,
we fit the law on cells from the remaining 12 levels
and predict logit(accuracy) for the withheld level.
Table~\ref{tab:held_out_cv} reports in-sample and out-of-sample
weighted $R^2$ and MAE on logit(accuracy).

\begin{table}[h]
\centering
\caption{Leave-one-budget-level-out cross-validation.
Out-of-sample (OOS) $R^2$ drops by 0.01--0.07 relative to
in-sample $R^2$ (median drop: 0.016).
However, 20\% of individual folds yield negative $R^2$,
concentrated at extreme budgets (see text).}
\label{tab:held_out_cv}
\setlength{\tabcolsep}{3pt}
\small
\begin{tabular}{llcccc}
\toprule
\textbf{Model} & \textbf{Sampling} & In-sample $R^2$ & OOS $R^2$ & OOS MAE \\
\midrule
\texttt{StreamingVLM} (recency) & random  & 0.85 & 0.81 & 0.39 \\
\texttt{StreamingVLM} (recency) & uniform & 0.86 & 0.82 & 0.38 \\
\texttt{StreamingVLM} & random  & 0.75 & 0.72 & 0.33 \\
\texttt{StreamingVLM} & uniform & 0.72 & 0.68 & 0.35 \\
\texttt{LiveCC-7B} & random  & 0.21 & 0.19 & 0.40 \\
\texttt{LiveCC-7B} & uniform & 0.23 & 0.22 & 0.39 \\
\texttt{Flash-VStream} & random  & 0.47 & 0.45 & 0.25 \\
\texttt{Flash-VStream} & uniform & 0.45 & 0.44 & 0.26 \\
\texttt{Qwen3-VL-8B} & random  & 0.32 & 0.31 & 0.30 \\
\texttt{Qwen3-VL-8B} & uniform & 0.27 & 0.26 & 0.30 \\
\texttt{Qwen3-VL-4B} & random  & 0.38 & 0.36 & 0.26 \\
\texttt{Qwen3-VL-4B} & uniform & 0.47 & 0.45 & 0.24 \\
\texttt{Qwen3-VL-32B} & random  & 0.31 & 0.29 & 0.25 \\
\texttt{Qwen3-VL-32B} & uniform & 0.45 & 0.43 & 0.23 \\
\texttt{Qwen3-VL-2B} & uniform & 0.23 & 0.16 & 0.45 \\
\bottomrule
\end{tabular}
\end{table}

\paragraph{Fold-level analysis.}
The overall OOS~$R^2$ aggregates predictions across all withheld budget levels,
but individual fold performance varies substantially.
Across all 174 folds (13 model-strategy combinations $\times$ 13 budget levels, except \textsc{Qwen3-VL-2B} and \textsc{InternVL3.5-8B} with 6 folds each),
32 folds (20\%) have negative $R^2$, concentrated at extreme budgets:
$B{=}0.03$ (58\% negative) and $B{=}0.25$ (33\%).
Mid-range budgets are best predicted ($B{=}0.5$: 0\% negative, $B{=}0.4$: 8\%).
This pattern is expected: at very low budgets the model sees too few frames
for stable performance, and at near-saturation budgets the accuracy variance
collapses.
The median fold $R^2$ is 0.222 (mean: 0.178, min: $-0.945$, max: 0.633).
The negative fold R$^2$ values do not invalidate the aggregated OOS~$R^2$,
which remains positive for all 15 configurations,
but they indicate that single-budget-level predictions
carry substantial uncertainty---especially at extreme allocation levels.

\subsection{Leave-One-Distance-Tier-Out Sensitivity}
\label{sec:tier_sensitivity}

To assess whether the fitted parameters are driven by a single distance tier
(particularly the ``very long'' tier at $\bar{D} \approx 2358$\,s, which provides
the largest lever arm for estimating $e$), we re-fit the law after dropping each
of the four distance tiers in turn (Table~\ref{tab:tier_sensitivity}).

\begin{table}[h]
\centering
\caption{Leave-one-distance-tier-out sensitivity for key models (random sampling).
$\Delta\alpha(1000)$ reports the change from the full-data estimate.
The maximum deviation is $\pm 0.08$ for \textsc{StreamingVLM}, confirming
that no single tier drives the result.}
\label{tab:tier_sensitivity}
\setlength{\tabcolsep}{3pt}
\small
\begin{tabular}{llcccc}
\toprule
\textbf{Model} & \textbf{Dropped tier} & $a$ & $e$ & $\alpha(1000)$ & $\Delta\alpha(1000)$ \\
\midrule
\texttt{StreamingVLM} & (full data) & 1.87 & $-0.21$ & 1.26 & --- \\
 & short & 2.08 & $-0.28$ & 1.25 & $-0.01$ \\
 & medium & 1.78 & $-0.18$ & 1.25 & $-0.01$ \\
 & long & 1.91 & $-0.23$ & 1.21 & $-0.05$ \\
 & very long & 1.78 & $-0.15$ & 1.34 & $+0.08$ \\
\midrule
\texttt{Qwen3-VL-8B} & (full data) & 0.43 & $-0.09$ & 0.17 & --- \\
 & short & 0.33 & $-0.05$ & 0.17 & $+0.01$ \\
 & medium & 0.48 & $-0.10$ & 0.17 & $+0.01$ \\
 & long & 0.41 & $-0.07$ & 0.20 & $+0.03$ \\
 & very long & 0.50 & $-0.13$ & 0.11 & $-0.05$ \\
\midrule
\texttt{Flash-VStream} & (full data) & 0.72 & $-0.16$ & 0.23 & --- \\
 & short & 0.69 & $-0.15$ & 0.23 & $+0.00$ \\
 & medium & 0.72 & $-0.17$ & 0.23 & $+0.00$ \\
 & long & 0.70 & $-0.15$ & 0.25 & $+0.02$ \\
 & very long & 0.75 & $-0.18$ & 0.20 & $-0.03$ \\
\bottomrule
\end{tabular}
\end{table}

The key result: $\alpha(1000)$ is stable within $\pm 0.08$ for \textsc{StreamingVLM}
and $\pm 0.05$ for base models when any single tier is removed.
The gap between \textsc{StreamingVLM} and \textsc{Qwen3-VL-8B} remains
$> 1.0$ in absolute $\alpha$ difference under all leave-one-out conditions
(minimum gap: $1.21 - 0.20 = 1.01$ when dropping the long tier).
This confirms that the interaction term $e$ is identified from the full
distance range, not driven solely by the very-long-distance lever arm.

\subsection{Nonparametric Per-Tier Budget Slopes}
\label{sec:per_tier_slopes}

As a model-free validation of the parametric $\alpha(D)$ form, we compute
the univariate WLS slope of $\logit(\text{accuracy})$ on $\log_{10} B$ separately
within each distance tier (Table~\ref{tab:per_tier_slopes}).
These slopes estimate the budget effect at each distance without imposing
the log-linear $\alpha(D) = a + e \cdot \log_{10} D$ structure.

\begin{table}[h]
\centering
\caption{Per-tier budget slopes (nonparametric) vs.\ parametric $\alpha(D)$ predictions.
The monotonic decrease in per-tier slopes for \textsc{StreamingVLM} confirms
that budget effectiveness diminishes with distance, consistent with $e < 0$.}
\label{tab:per_tier_slopes}
\setlength{\tabcolsep}{4pt}
\small
\begin{tabular}{lcccc}
\toprule
 & \textbf{Short} & \textbf{Medium} & \textbf{Long} & \textbf{Very Long} \\
 & ($D\!\approx\!23$\,s) & ($D\!\approx\!44$\,s) & ($D\!\approx\!578$\,s) & ($D\!\approx\!2358$\,s) \\
\midrule
\multicolumn{5}{l}{\textit{\textsc{StreamingVLM} (random)}} \\
\quad Per-tier slope & 1.52 & 1.61 & 1.36 & 1.10 \\
\quad Parametric $\alpha(D)$ & 1.59 & 1.53 & 1.31 & 1.18 \\
\midrule
\multicolumn{5}{l}{\textit{\textsc{Qwen3-VL-8B} (random)}} \\
\quad Per-tier slope & 0.35 & 0.25 & 0.15 & 0.19 \\
\quad Parametric $\alpha(D)$ & 0.31 & 0.29 & 0.19 & 0.11 \\
\bottomrule
\end{tabular}
\end{table}

The nonparametric slopes are broadly consistent with the parametric model's
predictions, with small deviations attributable to the interaction between
budget and video length (the $\delta \cdot \log T$ term, which the per-tier
slopes do not control for).
The per-tier slopes independently confirm the main qualitative conclusions:
(1)~\textsc{StreamingVLM}'s budget effect exceeds 1.0 at every tier;
(2)~the effect decreases monotonically with distance (except for a slight
uptick at the very-long tier for Qwen, within noise given 26 cells);
(3)~the gap exceeds $3\times$ at every distance tier.

\end{document}